\documentclass[letterpaper]{article} 
\usepackage[preprint]{aaai2027}  
\usepackage[hyphens]{url}  
\usepackage{graphicx} 
\usepackage{natbib}  
\usepackage{caption} 
\usepackage{algorithm}
\usepackage{algorithmic}
\usepackage{multirow}
\usepackage{amsmath}
\usepackage{amssymb}
\usepackage{newfloat}
\usepackage{listings}
\usepackage{booktabs}
\usepackage{multirow}
\usepackage{graphicx}
\DeclareCaptionStyle{ruled}{labelfont=normalfont,labelsep=colon,strut=off} 
\floatstyle{ruled}
\newfloat{listing}{tb}{lst}{}
\floatname{listing}{Listing}

\usepackage{booktabs}

\title{DeMixPert: Decomposed Response Modeling with Gaussian Mixtures for OOD Single-Cell Perturbation Prediction}

\author{
    Jiawen Liu\textsuperscript{\rm 1},
    Xuechenxiao Cao\textsuperscript{\rm 1},
    Yutong Li\textsuperscript{\rm 2},
    Bing Liu\textsuperscript{\rm 1},
    Jiaming Liang\textsuperscript{\rm 1},
    Tinghe Zhang\textsuperscript{\rm 2},
    Xiaoqi Sheng\textsuperscript{\rm 2},
    Hongmin Cai\textsuperscript{\rm 1,2}\corresponding
}
\affiliations{
    \textsuperscript{\rm 1}School of Computer Science and Engineering, South China University of Technology, Guangzhou, China\\
    \textsuperscript{\rm 2}School of Future Technology, South China University of Technology, Guangzhou, China\\
    \textsuperscript{*}hmcai@scut.edu.cn
}

\begin{document}

\maketitle

\begin{abstract}
Predicting transcriptome-wide responses to unseen genetic perturbations remains a major computational challenge because accurate prediction requires recovering both perturbation-specific transcriptional shifts and heterogeneous cellular responses.
Existing methods often entangle deterministic response structure with stochastic population-level variation, causing dominant shared patterns to mask weaker perturbation-specific signals and impair distributional modeling.
To address these challenges, we propose \textbf{DeMixPert}, an approach for \textbf{De}composed response Modeling with Gaussian \textbf{Mix}tures for Out-Of-Distribution (OOD) single-cell \textbf{Pert}urbation prediction.  DeMixPert decomposes perturbation-induced changes into a basal-state-dependent systematic response, a perturbation-specific response, and population-level variation.
The systematic component is derived from the basal state encoded from control-cell expression, whereas the perturbation-specific component is inferred from pretrained target embeddings for unseen-target generalization.
DeMixPert models population-level variation using a Gaussian prototype Invertible Network and adaptively combines reusable Gaussian prototypes according to the basal state and perturbation condition. The resulting mixture is mapped to a condition-specific variation distribution. 
Sampled variations are integrated with the systematic and perturbation-specific components, followed by joint decoding with the basal state to reconstruct perturbed-cell gene expression.
Experimental results show that DeMixPert effectively captures heterogeneous single-cell perturbation responses and achieves superior performance across unseen-perturbation settings. The source code is made publicly available upon publication.
\end{abstract}

\section{Introduction}
Single-cell genetic perturbation profiling couples controlled genetic interventions with transcriptome-wide readouts, enabling cell-resolved characterization of transcriptional responses and the regulatory programs underlying them~\cite{adamson2016multiplexed,dixit2016perturb}. 
Despite its value, the scalability of such profiling is limited by the combinatorial growth of experimental conditions across perturbation targets, cellular states, and biological contexts~\cite{cheng2026prescribe}, making exhaustive measurement infeasible.
Accordingly, \textit{in silico} perturbation-response prediction offers a scalable means of estimating cellular responses under experimentally profiled conditions.


\begin{figure}[!t]
    \centering
    \includegraphics[
        width=0.8\columnwidth,
        height = 0.19\textheight
    ]{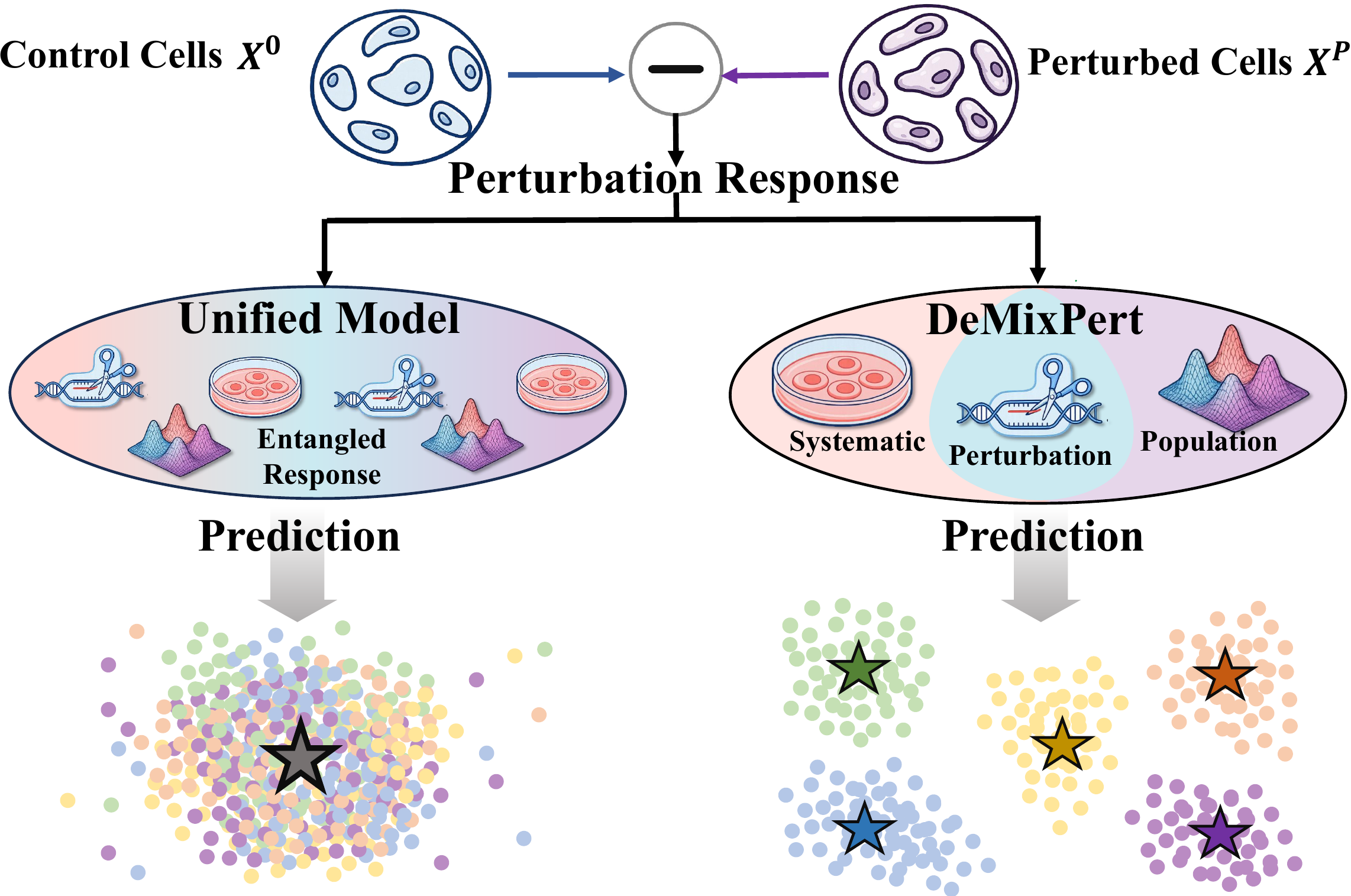}
\caption{Comparison of unified and decomposed response modeling. By separating entangled response components, DeMixPert better preserves perturbation-specific signals and condition-dependent population structure.}
    \label{fig:res}
\end{figure}
 However, predicting responses under experimentally unobserved conditions involves more than estimating an average perturbation-induced expression shift~\cite{yu2025perturbnet}.
 These challenges can be understood through three intertwined components of perturbation responses, as illustrated in Fig.~\ref{fig:res}.
 First, the basal cellular state contributes a systematic component that shapes responses across perturbation conditions~\cite{dong2023causal, song2025decoding}.  This shared structure should be learned from observed conditions and transferred to unseen ones.
 Second, each perturbation condition induces a perturbation-specific transcriptional effect that distinguishes its response from those of other interventions~\cite{norman2019exploring,replogle2020}. 
For an unseen perturbation, this effect must be inferred from biological relationships with targets observed during training.
Third, cells exposed to the same perturbation may occupy distinct response subpopulations,
whose internal structure and relative abundance vary with cellular state and environmental context~\cite{adamson2016multiplexed, frangieh2021multimodal}. Consequently, Out-Of-Distribution~(OOD) prediction imposes three distinct requirements involving systematic-structure transfer, unseen-target effect inference, and condition-specific population-distribution estimation without direct observations.


Existing methods improve perturbation-response prediction from different perspectives. Latent-variable approaches represent perturbation responses through state transformations or factorized perturbation components~\cite{lotfollahi2023predicting}. In parallel, Knowledge-guided methods exploit gene representations or biological priors to generalize from limited perturbation observations~\cite{cui2024scgpt}. From another perspective, population-distribution approaches learn transitions from control to perturbed populations using optimal transport, set-level distributional objectives, or flow matching~\cite{bunne2023learning,adduri2025predicting,yu2026scdfm}. Despite substantial advances in response transfer and distributional modeling, most approaches still formulate the full perturbation response as a monolithic prediction target, leaving shared systematic response, perturbation-specific response, and condition-dependent population-level variation entangled.
This limitation is consequential because Systema shows that shared systematic variation can dominate standard evaluation metrics~\cite{vinas2025systema}. Accordingly, accurate OOD prediction requires disentangling basal-state-dependent systematic responses, perturbation-specific responses, and population-level variation.


In this paper, we propose \textbf{DeMixPert}, a \textbf{De}composed
response framework with Gaussian \textbf{Mix}tures for
out-of-distribution single-cell \textbf{Pert}urbation prediction. By decomposing perturbation responses, DeMixPert assigns shared systematic response, perturbation-specific response, and population
variation to separate modules. The basal-state-dependent systematic response is integrated with the perturbation-specific response to define a deterministic response center. This separation preserves transferable response structure while preventing shared variation from masking perturbation-specific signals.
Subsequently, a Gaussian prototype Invertible Network models population-level variation around the response center by adaptively weighting reusable Gaussian prototypes according to the basal state and perturbation
embedding. The resulting mixture is transformed into a
condition-specific variation distribution through an invertible
mapping. Finally, sampled variations are added to the response center and decoded together with the basal state to reconstruct
post-perturbation gene expression. This design improves response recovery for unseen perturbation targets while preserving perturbation discriminability and distributional fidelity. Our main contributions are summarized as follows:
\begin{itemize}
\item We propose DeMixPert, a decomposed response model for OOD single-cell perturbation prediction. DeMixPert disentangles deterministic response structure from stochastic cellular heterogeneity and captures condition-specific response distributions using adaptive Gaussian mixtures.

\item We introduce a Gaussian prototype Invertible Network for population-level variation modeling. The invertible network uses context-adaptive weights to combine reusable Gaussian mixture prototypes, thereby estimating condition-specific population-variation distributions.

\item Quantitative analyses demonstrate that DeMixPert improves OOD perturbation prediction by accurately recovering perturbation-specific responses while preserving population-level distributional fidelity.

\end{itemize}

\begin{figure*}[htb]
	\centering
	\includegraphics[width=1.0\linewidth, height = 0.3\textheight]{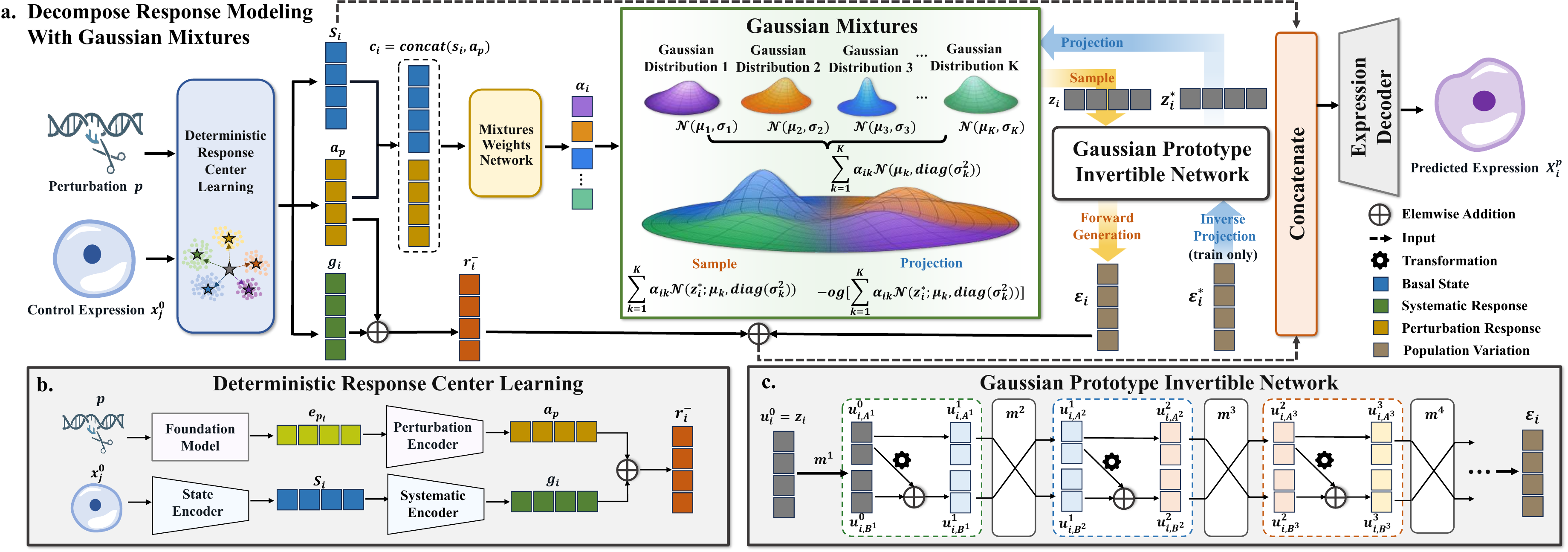}
	\caption{Overview of DeMixPert. \textbf{(a)} The overall workflow of DeMixPert. \textbf{(b)} The architecture of the Deterministic Response Center Learning module. \textbf{(c)} The architecture of the Gaussian prototype Invertible Network.}
    \label{fig1}
\end{figure*}

\section{Related Work}
\subsection{Genetic Perturbation Prediction}
Existing approaches formulate cellular responses as latent transformations: scGen applies an average shift, whereas CPA factorizes treatment, dose, and covariate contributions \cite{lotfollahi2019scgen,lotfollahi2023predicting}. For unseen targets, GEARS leverages gene-relation graphs, while GenePert draws on external gene embeddings \cite{roohani2024predicting,chen2024genepert}. To capture heterogeneity in unpaired data, STATE predicts post-intervention cell sets, whereas scDFM learns full conditional distributions via flow matching \cite{adduri2025predicting,yu2026scdfm}. Systema further reveals that shared systematic variation can dominate standard metrics and mask target-specific signals \cite{vinas2025systema}. Nevertheless, most prior work does not jointly disentangle basal-state-dependent systematic responses, perturbation-specific responses, and condition-dependent population-level variation.

\subsection{Gaussian Mixtures and Distribution Modeling}

Gaussian mixture (GM) models provide a flexible representation of complex population structure \cite{mclachlan2000finite}, while normalizing flows transform tractable source distributions through invertible mappings \cite{dinh2014nice}. PerturbNet combines perturbation embeddings with a conditional invertible network to generate cell-state distributions \cite{yu2025perturbnet}. More recently, MixFlow replaces the conventional unimodal Gaussian source with a descriptor-conditioned Gaussian mixture and transports it using conditional flow matching \cite{rubbi2026mixflow}.
Despite improved distributional flexibility, existing approaches still model perturbation responses monolithically, conflating systematic and perturbation-specific effects with residual variation. This entanglement limits component-wise generalization and mechanistic interpretation.

\section{Methodology}
\subsection{Preliminaries}
In the context of OOD single-cell perturbation prediction, we decompose responses into systematic, perturbation-specific, and population-level components to capture transferable patterns and condition-specific heterogeneity.

Let $
\mathcal{D}
=
\left\{
X^0,\{X^p\}_{p\in\mathcal{P}}
\right\}
$
denote a single-cell genetic perturbation dataset, where
$\mathcal{P}$ is the set of perturbation conditions, each involving
one or more target genes. The control population is defined by
$
X^0=\{x_j^0\in\mathbb{R}^{G}\}_{j=1}^{N_0}.
$
Here, $N_0$ is the number of control cells and $G$ denotes the number of selected
genes. 
For each perturbation condition $p\in\mathcal{P}$, the
corresponding perturbed population is defined by
$
X^p=\{x_i^p\}_{i=1}^{N_p},
$
where $N_p$ is the number of cells observed under perturbation $p$.

Since the control and perturbed populations are unpaired, for each perturbed cell $x_i^p$, we randomly sample a control cell $x_j^0$ from $X^0$ as its reference and define the perturbation response as
$
\Delta x_i^p = x_i^p - x_j^0.
$
To facilitate subsequent response modeling, the expression-space response $\Delta x_i^p$ is projected into a latent space using an encoder $E_r$: $
r_i^*
=
E_r(\Delta x_i^p)
\in
\mathbb{R}^{d_r}
$. The objective of DeMixPert is to predict a perturbation response $\hat{r}_i$ that approximates the target response $r_i^*$. In DeMixPert, $\hat{r}_i$ is modeled as three
components: $
\hat{r}_i
=
g_i+a_p+\epsilon_i.
$
Here, $g_i$, $a_p$, and $\epsilon_i$ denote the systematic response, perturbation-specific response, and population-level response variation, respectively.

\subsection{Overview}
Fig.~\ref{fig1} outlines the DeMixPert framework. As shown in Fig.~\ref{fig1}~(a), the Deterministic Response Center Learning module takes the perturbation condition and control-cell expression as inputs. The module separates the response into systematic and perturbation-specific components, whose combination forms the deterministic response center. Meanwhile, the basal-state and perturbation representations are concatenated to predict mixture weights, adapting a set of shared Gaussian prototypes to the current cell state and perturbation condition.
The adapted Gaussian mixture is connected to the Gaussian prototype Invertible Network for population-variation modeling. The forward transformation generates population-level variations from source samples. Conversely, the training-only inverse transformation projects observed target variations into the Gaussian-mixture space. The corresponding conditional likelihood provides joint supervision for mixture-weight estimation and invertible transformation learning. Generated population-level variations are subsequently added to the deterministic response center. Thereafter, the resulting response representation is concatenated with the basal-state representation for expression decoding.

\subsection{Deterministic Response Center Learning}

To prevent systematic variation from obscuring perturbation-specific
effects, DeMixPert constructs a deterministic response center by combining separately learned systematic and perturbation-specific responses.
The three encoders $E_s$, $E_{\mathrm{sys}}$, and $E_{\mathrm{pert}}$ produce the basal-state representation, systematic response, and perturbation-specific response, respectively: $s_i=E_s(x_j^0)\in\mathbb{R}^{d_s}$, $g_i=E_{\mathrm{sys}}(s_i)\in\mathbb{R}^{d_r}$, and $a_p=E_{\mathrm{pert}}(e_p)\in\mathbb{R}^{d_r}$.
Here, $s_i$ summarizes the unperturbed cellular context and is
retained for subsequent variation modeling. The systematic response $g_i$ captures basal-state-dependent patterns shared across perturbation conditions, whereas $a_p$ captures the perturbation-specific response derived from the perturbation embedding $e_p$.
We derive $e_p$ from pretrained scGPT representations~\cite{cui2024scgpt} for OOD generalization. 
The $e_p$ provides prior knowledge of biological relationships among genes for inferring perturbation-specific responses without direct observations.

Finally, the resulting systematic and
perturbation-specific responses are combined to form the deterministic response center: $
\bar{r}_i
=
g_i+a_p
\in
\mathbb{R}^{d_r}.$
The $\bar{r}_i$ denotes the deterministic response pattern
jointly specified by the basal cellular state $s_i$ and perturbation target $e_p$ in the response latent space $\mathbb{R}^{d_r}$.


\subsection{Gaussian prototype Invertible Network}
The deterministic response center captures systematic and
perturbation-specific responses but not the conditional
population-level variation distribution. Accordingly, DeMixPert models the remaining population-level variation using a Gaussian prototype Invertible Network. 
Using the target response $r_i^*$, the
target variation from the deterministic response center is given by $
    \epsilon_i^*
    =
    r_i^*-\bar{r}_i
    \in\mathbb{R}^{d_r}.$

For each perturbation condition, the target variations of its observed cells form samples from the corresponding population-variation
distribution. To capture potential variation patterns
shared across perturbations, the target variations
$\epsilon^*$ pooled from all training conditions are modeled using a $K$-component diagonal Gaussian mixtures:
\begin{equation}
    q_{\mathrm{GM}}(\epsilon)
    =
    \sum_{k=1}^{K}
    \pi_k
    \mathcal{N}\left(
        \epsilon;
        \mu_k,
        \operatorname{diag}(\sigma_k^2)
    \right),
    \label{eq:global_GM}
\end{equation}
where $\pi_k$, $\mu_k$, and $\operatorname{diag}(\sigma_k^2)$ denote the weight, mean, and diagonal variance of the $k$-th Gaussian prototype, respectively.
These parameters are periodically re-estimated from the target
variations using the expectation-maximization algorithm
~\cite{dempster1977maximum}. 
Further details are provided in Supplementary Appendix~F.

The global Gaussian Mixtures provide reusable Gaussian prototypes shared across perturbations. However, the contribution of each prototype depends
jointly on the basal state and perturbation condition. Accordingly, $s_i$ and $a_p$ are concatenated to predicts Gaussian prototype weights:
\begin{equation}
\alpha_i=
\operatorname{softmax}\left(\operatorname{MLP}(\operatorname{concat}(s_i,a_p))\right)
\in\mathbb{R}^{K},
\label{eq:adaptive_prototype_weights}
\end{equation}
where $\operatorname{concat}(\cdot,\cdot)$ denotes concatenation, $\operatorname{MLP}(\cdot)$ denotes a multilayer perceptron, and $\operatorname{softmax}(\cdot)$ normalizes the predicted prototype weights.
The conditional Gaussian mixture is then defined as:
\begin{equation}
    P(z\mid \alpha_{i})
    =
    \sum_{k=1}^{K}
    \alpha_{ik}
    \mathcal{N}\left(
        z;
        \mu_k,
        \operatorname{diag}(\sigma_k^2)
    \right).
    \label{eq:conditional_source}
\end{equation}

The prototype parameters are shared globally, while their contributions are
adapted to the condition representation $c_i$.
Then, a source variable $z_i\sim P(z\mid \alpha_i)$ is sampled from the
conditional mixture and passed through the invertible network $T_\theta$,
producing a sample ${\epsilon}_i$ from the final population-variation 
distribution: $ \epsilon_i=T_\theta(z_i;c_i).$

Let $u_i^{(0)}=z_i$ and 
$
u_i^{l}
=
F_{l}\!\left(u_i^{(l-1)};c_i\right)
\in\mathbb{R}^{d_r}
$
denote the output of the $l$-th coupling layer.  A binary mask $m_{l}\in\{0,1\}^{d_r}$ partitions the feature dimensions into two complementary index sets, $A^{l}$ and $B^{l}$, corresponding to mask values of $1$ and $0$, respectively, such that
$
A^{\ell}\cap B^{\ell}=\emptyset$, $
A^{\ell}\cup B^{\ell}=\{1,\ldots,d_r\}.
$
Accordingly, the two input subvectors are defined as
$
u_{i,A^{l}}^{(l-1)}
:=
\bigl(u_{ij}^{(l-1)}\bigr)_{\in A^{l}}
\in\mathbb{R}^{|A^{l}|}$, 
$
u_{i,B^{l}}^{(l-1)}
:=
\bigl(u_{ij}^{(l-1)}\bigr)_{j\in B^{l}}
\in\mathbb{R}^{|B^{l}|}.
$ The $u_i^{(l-1)}$ can be recovered by restoring the two
subvectors to the positions specified by $m_{l}$:
$
u_i^{(l-1)}
=
\operatorname{merge}_{m_{l}}\!\left(
u_{i,A^{l}}^{(l-1)},
u_{i,B^{l}}^{(l-1)}
\right).
$
Here, $\operatorname{merge}_{m^l}$ restores the two subsets to their original feature positions specified by $m^l$. The mask is fixed within each layer but varies across layers, allowing all feature dimensions to be updated.


Within the $l$-th coupling layer, 
$u_{i,A^{l}}^{(l-1)}$ remain unchanged and is used to determine the transformation applied to $u_{i,B^{l}}^{(l-1)}$. DeMixPert
first measures how closely the unchanged features match each Gaussian
prototype. The normalized matching weight associated with prototype
$k$ is
\begin{equation}
    \gamma_{ik}^{l}
=
\frac{
\pi_k
\mathcal{N}\!\left(
u_{i,A^{l}}^{(l-1)};
\mu_{k,A^{l}},
\operatorname{diag}\!\left(\sigma_{k,A^{l}}^{2}\right)
\right)
}{
\displaystyle
\sum_{q=1}^{K}
\pi_q
\mathcal{N}\!\left(
u_{i,A^{l}}^{(l-1)};
\mu_{q,A^{l}},
\operatorname{diag}\!\left(\sigma_{q,A^{l}}^{2}\right)
\right)
}.
\end{equation}

In parallel, a layer-specific MLP $h^{l}$ produces prototype modulation coefficients from $c_i$:
$
    b_i^{l}
    =
    h^{l}(c_i)
    \in\mathbb{R}^{K}.
$ 
Here, $\gamma_i^{l}$ captures feature-level prototype matching, whereas
$b_i^{l}$ captures the contribution of each prototype under the
current basal-state and perturbation context.
The two coefficients are combined to generate the additive shift applied
to $B^l$:
\begin{equation}
    f^{l}(u_{i,A^l},c_{i})
    =
    W^l
    \left(
        \gamma_i^{l}
        \odot
        b_i^{l}
    \right)
    \in\mathbb{R}^{|B^l|},
    \label{eq:prototype_shift}
\end{equation}
where $W^l\in\mathbb{R}^{|B^l|\times K}$ is a learnable projection
matrix and $\odot$ denotes element-wise multiplication. The additive coupling transformation is defined as follows: 
$
    u_{i,A^l}^{l}
    =
    u_{i,A^l}^{l-1},\quad
    u_{i,B^l}^{l}
    =
    u_{i,B^l}^{l-1}
    +
    f_{l}
    \left(
        u_{i,A^l}^{l-1},c_i
    \right).$
The two subsets are then reassembled to obtain the complete output of the
$l$-th coupling layer:
\begin{equation}
        F_l\left(u_i^{l-1};c_i\right)=u_i^{l}
    =
    \operatorname{merge}_{m^l}
    \left(
        u_{i,A^l}^{l},
        u_{i,B^l}^{l}
    \right)
    .
\end{equation}

Since the $A^l$ subset remains unchanged, the same shift function can be evaluated during inversion \cite{dinh2016density}. The inverse transformation is defined by $u_{i,A^l}^{l-1}
    =
    u_{i,A^l}^{l}$ and
    $u_{i,B^l}^{l-1}
    =
    u_{i,B^l}^{l}
    -
    f_{l}
    \left(
        u_{i,A^l}^{l},c_i
    \right).$

The recovered subsets are reassembled to obtain the complete output of the
inverse layer:
\begin{equation}
    F_l^{-1}\left(u_i^{l};c_i\right)=u_i^{l-1}
    =
    \operatorname{merge}_{m^l}
    \left(
        u_{i,A^l}^{l-1},
        u_{i,B^l}^{l-1}
    \right).
    \label{eq:inverse_coupling_output}
\end{equation}

Each coupling layer is bijective and has a triangular Jacobian
with unit determinant~\cite{dinh2014nice}. 
This invertibility maps each target variation  back to the Gaussian mixtures. Its likelihood under the source distribution supervises the
context-dependent prototype weights and the invertible transformation. The
corresponding likelihood objective is introduced in the training loss.

After $L$ coupling layers, the generated variation  sample is
\begin{equation}
  \epsilon_i
    =
    T_\theta(z_i;c_i)
    =
    F_L\circ\cdots\circ F_1(z_i;c_i)
    \in\mathbb{R}^{d_r}.
    \label{eq:generated_variation }
\end{equation}

For perturbation condition $p$, the collection
$\{\epsilon_i:p_i=p\}$ forms samples from its predicted
population-level variation distribution. Each sample perturbs the deterministic response center to yield the response latent: $\hat{r}_i = \bar{r}_i+\epsilon_i.$

\subsection{Expression Decoder}
The post-perturbation expression is jointly determined by the original cellular context and the perturbation-induced response. The basal-state
representation $s_i$ captures the cellular context, whereas the predicted response
latent $\hat{r}_i$ encodes the perturbation effect. The concatenated representation is then decoded to reconstruct post-perturbation gene expression:
$\hat{x}_i^{p}
    =
    D_{\psi}\left(\operatorname{concat}(s_i,\hat{r}_i)\right).$
Therein, $D_{\psi}$ denotes the expression decoder.

\subsection{Training Objective}


DeMixPert is trained end-to-end to jointly learn the deterministic response center and condition-specific population-level variation. The total training objective consists of four loss components that supervise response alignment, population-variation likelihood, gene-expression reconstruction, and population-distribution matching.

\noindent\textbf{Response Alignment Loss.}
A mean squared error loss is used to align the predicted response latent with its target:
$
\mathcal{L}_{\mathrm{align}}
=
\frac{1}{N}
\sum_{i=1}^{N}
\left\|
\hat{r}_i-r_i^{*}
\right\|_2^2,
$
where $N$ denotes the number of cells in the training batch.

\noindent\textbf{Population-Variation Likelihood Loss.}
To learn the conditional population-level variation, each target variation is first mapped back to the Gaussian mixtures:
$
z_i^{*}
=
T_{\theta}^{-1}(\epsilon_i^{*};c_i).
$
We then minimize its negative log-likelihood under the conditional Gaussian mixtures:
\begin{equation}
\mathcal{L}_{\mathrm{gm}}
=
-\frac{1}{N}
\sum_{i=1}^{N}
\log
\left[
\sum_{k=1}^{K}
\alpha_{ik}
\mathcal{N}\left(
z_i^{*};
\mu_k,
\operatorname{diag}(\sigma_k^2)
\right)
\right],
\end{equation}
where, the $\mathcal{L}_{\mathrm{gm}}$ jointly supervises the context-dependent prototype weights and the invertible transformation.

\noindent\textbf{Gene-Expression Reconstruction Loss.}
The reconstruction loss preserves gene expression information by reconstructing the perturbed expression from both the predicted and target responses, while recovering the control expression from a zero-response vector $\mathbf{0}\in\mathbb{R}^{d_r}$:
\begin{equation}
\begin{aligned}
\mathcal{L}_{\mathrm{rec}}
={}&
\mathcal{L}_{mse}\!\left(
D_{\psi}([s_i,\hat{r}_i]),x_i^{p_i}
\right)
+
\mathcal{L}_{mse}\!\left(
D_{\psi}([s_i,r_i^{*}]),x_i^{p_i}
\right)
\\
&+
\mathcal{L}_{mse}\!\left(
D_{\psi}([s_i,\mathbf{0}]),x_j^{0}
\right),
\end{aligned}
\end{equation}
where $\mathcal{L}_{mse}$ denotes the mean square error loss.

\noindent\textbf{Distribution Loss.}
At the population level, we use energy distance to align the generated and observed expression distributions under each perturbation condition:

\begin{equation}
    \mathcal{L}_{\mathrm{dist}}
=
\frac{1}{|\mathcal{P}|}
\sum_{p\in\mathcal{P}}
\operatorname{ED}\left(
\{\hat{x}_i^{p}\}_{i:p_i=p},
\{x_i^{p}\}_{i:p_i=p}
\right),
\end{equation}
where $\mathcal{P}$ denotes the set of perturbation conditions and $\operatorname{ED}(\cdot)$ is the energy distance~\cite{szekely2013energy}.
The overall training objective is
$
    \mathcal{L}
=
\mathcal{L}_{\mathrm{align}}
+
\lambda_{\mathrm{gm}}\mathcal{L}_{\mathrm{gm}}
+
\mathcal{L}_{\mathrm{rec}}
+
\mathcal{L}_{\mathrm{dist}}.
$
The $\lambda_{\mathrm{gm}}$ balances the numerical scale of the negative log-likelihood against the other loss components. All trainable parameters are jointly optimized by minimizing $\mathcal{L}$.

\section{Experiments}
\subsection{Experimental Setup}
\paragraph{Datasets and preprocessing.}
We benchmark DeMixPert on four widely used single-cell genetic
perturbation datasets. Adamson \cite{adamson2016multiplexed} and Papalexi \cite{papalexi2021} contain single-gene perturbations, whereas Norman \cite{norman2019exploring} and Replogle \cite{replogle2020} involve combinatorial perturbations. All datasets are processed following a standard single-cell RNA-seq preprocessing pipeline. For each dataset, we select 2,048 highly variable genes (HVGs) and additionally retain all perturbation-target genes to define
the input gene space. 
Detailed dataset descriptions and statistics are provided in Appendix~C.
\paragraph{Compared methods.}
We compare DeMixPert against five representative perturbation-response prediction methods: GEARS \cite{roohani2024predicting}, scGPT
\cite{cui2024scgpt}, GenePert \cite{chen2024genepert}, scDFM
\cite{yu2026scdfm}, and STATE \cite{adduri2025predicting}. 
These methods span graph learning, pretrained foundation models, ridge regression, conditional flow matching, and set-based Transformers.
Accordingly, the comparative experiments enable DeMixPert to be evaluated against methods with substantially different assumptions and predictive mechanisms.

\paragraph{Metrics and implementation.}
Model performance is evaluated at both the top-100 differentially
expressed genes (DEGs) level and the all-gene level. At the top-100
DEG level, we report Common Differentially Expressed Genes (C-DEGs),
Energy Distance (E-Dist), Wasserstein Distance (W-Dist), and Mean
Squared Error (MSE). At the all-gene level, we report the Differential
Expression Score (DES), MSE, Centroid Accuracy (Centroid Acc), and the
Perturbation Discrimination Score (PDS)
\cite{wei2026benchmarking,vinas2025systema,roohani2025virtual}.
Together, these metrics assess differential-expression recovery,
distributional agreement, expression error, and perturbation
discrimination.

Condition-disjoint splits evaluate unseen perturbations. Single-gene
datasets use 70\%/10\%/20\% training/validation/test splits. For
combinatorial datasets, training includes all single-gene conditions
and half of the combinatorial conditions, while the remainder is
divided equally between validation and test sets. 
Validation selects
checkpoints, and results are reported as mean $\pm$ standard variation across random seeds. DeMixPert uses PyTorch and AdamW.
Default and dataset-specific configurations are provided in
AppendiX~B and~H, respectively.

\providecommand{\val}[2]{%
  #1\,{\scriptsize $\pm$\,#2}%
}

\providecommand{\bestval}[2]{%
  {\bfseries #1\,{\scriptsize\boldmath $\pm$\,#2}}%
}

\providecommand{\vcenterhead}[1]{%
  \raisebox{0.75ex}[0pt][0pt]{#1}%
}

\providecommand{\tworowhead}[1]{%
  \multirow[c]{2}{*}{\raisebox{-0.75ex}{\textbf{#1}}}%
}

\begin{table*}[t]
\centering
\begingroup
\footnotesize
\renewcommand{\arraystretch}{0.92}
\setlength{\tabcolsep}{2.5pt}

\begin{tabular}{@{}llcccccccc@{}}
\toprule

\tworowhead{Dataset}
& \tworowhead{Method}
& \multicolumn{4}{c}{\textbf{Top 100 DEGs Level}}
& \multicolumn{4}{c}{\textbf{All Gene Level}} \\

\cmidrule(lr){3-6}
\cmidrule(lr){7-10}

&
& \vcenterhead{\textbf{C-DEGs}$\uparrow$}
& \vcenterhead{\textbf{E-Dist}$\downarrow$}
& \vcenterhead{\textbf{W-Dist}$\downarrow$}
& \vcenterhead{\textbf{MSE}$\downarrow$}
& \vcenterhead{\textbf{DES}$\uparrow$}
& \vcenterhead{\textbf{MSE}$\downarrow$}
& \shortstack[c]{\textbf{Centroid}\\[-0.2ex]\textbf{Acc}$\uparrow$}
& \vcenterhead{\textbf{PDS}$\uparrow$} \\

\midrule

\multirow{6}{*}{\textbf{Papalexi}}
& GEARS
& \val{2.280}{0.460}
& \val{0.617}{0.218}
& \val{12.720}{0.263}
& \val{0.065}{0.032}
& \val{0.194}{0.115}
& \val{0.024}{0.015}
& \val{0.240}{0.089}
& \val{0.640}{0.049} \\

& scGPT
& \val{1.040}{0.862}
& \val{0.208}{0.327}
& \bestval{3.468}{1.159}
& \val{0.056}{0.059}
& \val{0.077}{0.045}
& \val{0.010}{0.011}
& \val{0.500}{0.000}
& \val{0.623}{0.024} \\

& GenePert
& \val{5.427}{0.889}
& \val{0.337}{0.136}
& \val{7.011}{0.190}
& \val{0.027}{0.012}
& \val{0.127}{0.067}
& \val{0.004}{0.002}
& \val{0.500}{0.000}
& \val{0.600}{0.025} \\

& scDFM
& \val{5.560}{1.506}
& \val{1.971}{0.582}
& \val{14.053}{0.324}
& \val{0.274}{0.104}
& \bestval{0.212}{0.099}
& \val{0.099}{0.037}
& \val{0.200}{0.141}
& \val{0.568}{0.095} \\

& STATE
& \val{4.760}{1.499}
& \val{0.760}{0.214}
& \val{13.364}{0.222}
& \val{0.072}{0.033}
& \val{0.198}{0.107}
& \val{0.025}{0.014}
& \val{0.160}{0.089}
& \val{0.600}{0.000} \\

& \textbf{DeMixPert}
& \bestval{22.000}{4.854}
& \bestval{0.171}{0.088}
& \val{6.369}{0.403}
& \bestval{0.008}{0.006}
& \val{0.200}{0.129}
& \bestval{0.002}{0.001}
& \bestval{0.720}{0.110}
& \bestval{0.904}{0.046} \\

\midrule

\multirow{6}{*}{\textbf{Adamson}}
& GEARS
& \val{2.280}{0.145}
& \val{0.563}{0.125}
& \val{9.654}{0.217}
& \val{0.046}{0.015}
& \val{0.313}{0.082}
& \val{0.012}{0.003}
& \val{0.133}{0.067}
& \val{0.605}{0.026} \\

& scGPT
& \val{4.151}{0.131}
& \val{2.997}{0.310}
& \bestval{4.936}{0.326}
& \val{0.037}{0.014}
& \bestval{0.378}{0.078}
& \val{0.005}{0.002}
& \bestval{0.586}{0.079}
& \val{0.567}{0.037} \\

& GenePert
& \val{0.063}{0.010}
& \val{0.468}{0.174}
& \val{4.938}{0.301}
& \val{0.031}{0.013}
& \val{0.105}{0.038}
& \bestval{0.002}{0.001}
& \val{0.533}{0.014}
& \val{0.570}{0.025} \\

& scDFM
& \val{3.400}{1.381}
& \val{1.355}{0.216}
& \val{10.771}{0.628}
& \val{0.132}{0.027}
& \val{0.278}{0.082}
& \val{0.048}{0.007}
& \val{0.067}{0.047}
& \val{0.539}{0.040} \\

& STATE
& \val{2.947}{0.357}
& \val{0.820}{0.138}
& \val{10.989}{0.169}
& \val{0.052}{0.017}
& \val{0.280}{0.091}
& \val{0.015}{0.003}
& \val{0.080}{0.030}
& \val{0.534}{0.004} \\

& \textbf{DeMixPert}
& \bestval{10.733}{5.198}
& \bestval{0.448}{0.052}
& \val{5.481}{0.113}
& \bestval{0.021}{0.003}
& \val{0.153}{0.078}
& \val{0.006}{0.001}
& \val{0.560}{0.076}
& \bestval{0.947}{0.020} \\

\midrule

\multirow{6}{*}{\textbf{Norman}}
& GEARS
& \val{3.850}{0.557}
& \val{0.855}{0.220}
& \val{6.380}{0.234}
& \val{0.049}{0.019}
& \val{0.334}{0.029}
& \val{0.008}{0.002}
& \val{0.237}{0.112}
& \val{0.802}{0.090} \\

& scGPT
& \val{6.644}{0.627}
& \val{3.174}{0.157}
& \bestval{5.627}{0.239}
& \val{0.028}{0.007}
& \bestval{0.540}{0.031}
& \val{0.004}{0.001}
& \bestval{0.940}{0.013}
& \val{0.886}{0.025} \\

& GenePert
& \val{19.181}{0.658}
& \val{1.113}{0.154}
& \val{6.305}{0.256}
& \val{0.078}{0.013}
& \val{0.404}{0.022}
& \val{0.008}{0.011}
& \val{0.639}{0.034}
& \val{0.646}{0.033} \\

& scDFM
& \val{15.350}{3.548}
& \val{0.711}{0.148}
& \val{7.607}{0.464}
& \val{0.046}{0.011}
& \val{0.389}{0.029}
& \val{0.009}{0.003}
& \val{0.463}{0.116}
& \val{0.890}{0.037} \\

& STATE
& \val{21.006}{5.872}
& \val{1.341}{0.217}
& \val{7.175}{0.177}
& \val{0.105}{0.021}
& \val{0.354}{0.013}
& \val{0.011}{0.002}
& \val{0.031}{0.000}
& \val{0.524}{0.014} \\

& \textbf{DeMixPert}
& \bestval{31.650}{11.739}
& \bestval{0.374}{0.056}
& \val{6.606}{0.269}
& \bestval{0.003}{0.0002}
& \val{0.506}{0.063}
& \bestval{0.003}{0.0002}
& \val{0.750}{0.163}
& \bestval{0.981}{0.005} \\

\midrule

\multirow{6}{*}{\textbf{Replogle}}
& GEARS
& \val{2.111}{0.176}
& \val{0.448}{0.060}
& \val{12.743}{0.114}
& \val{0.038}{0.009}
& \val{0.189}{0.047}
& \val{0.012}{0.002}
& \val{0.267}{0.127}
& \val{0.669}{0.058} \\

& scGPT
& \val{6.089}{0.755}
& \val{2.263}{0.280}
& \bestval{4.302}{0.483}
& \val{0.011}{0.004}
& \bestval{0.269}{0.060}
& \val{0.002}{0.001}
& \bestval{0.778}{0.125}
& \val{0.654}{0.087} \\

& GenePert
& \val{12.533}{2.881}
& \val{0.200}{0.047}
& \val{4.600}{0.550}
& \bestval{0.002}{0.001}
& \val{0.200}{0.047}
& \val{0.002}{0.001}
& \val{0.558}{0.048}
& \val{0.580}{0.008} \\

& scDFM
& \val{3.422}{1.630}
& \val{0.802}{0.091}
& \val{13.234}{0.272}
& \val{0.092}{0.015}
& \val{0.202}{0.053}
& \val{0.026}{0.004}
& \val{0.178}{0.061}
& \val{0.588}{0.052} \\

& STATE
& \bestval{12.711}{3.278}
& \val{0.510}{0.075}
& \val{13.008}{0.107}
& \val{0.040}{0.012}
& \val{0.207}{0.041}
& \val{0.011}{0.002}
& \val{0.111}{0.000}
& \val{0.548}{0.014} \\

& \textbf{DeMixPert}
& \val{12.378}{4.071}
& \bestval{0.155}{0.040}
& \val{5.614}{0.156}
& \val{0.005}{0.001}
& \val{0.227}{0.066}
& \bestval{0.002}{0.000}
& \val{0.778}{0.193}
& \bestval{0.938}{0.039} \\

\bottomrule
\end{tabular}

\endgroup
\caption{Comprehensive evaluation results on four genetic perturbation
datasets, reported as mean $\pm$ standard variation.
$\uparrow$ ($\downarrow$) indicates that higher (lower) values are better.
The best result for each metric is marked in \textbf{bold}.}
\label{table_merged_final_DeMixPert}
\end{table*}

\subsection{Quantitative Comparison}
Table~\ref{table_merged_final_DeMixPert} summarizes the results for unseen single-gene and combinatorial perturbations. On the single-gene datasets Papalexi and Adamson, DeMixPert substantially improves perturbation-specific and distributional recovery, achieving C-DEGs scores of 22.000 and 10.733, PDS values of 0.904 and 0.947, and the lowest E-Dist values of 0.171 and 0.448, respectively. Although ScDFM obtains the highest DES on Papalexi, its lower C-DEGs and PDS suggest that matching response magnitude alone is insufficient to recover perturbation-specific effects. From another direction, performance on unseen combinatorial perturbations evaluates the ability to preserve interaction-specific effects and generalize to novel gene combinations. On Norman and Replogle, DeMixPert achieves the highest PDS (0.981 and 0.938) and the lowest E-Dist (0.374 and 0.155). Its C-DEGs score reaches 31.650 on Norman, surpassing STATE’s 21.006, and 12.378 on Replogle, close to the best baseline result of 12.711. Moreover, scGPT’s high Centroid Acc of 0.940 but worst E-Dist of 3.174 on Norman demonstrates that accurate centroid prediction does not guarantee distributional recovery. Overall, these results show that DeMixPert generalizes effectively across both perturbation regimes while preserving perturbation specificity and population-level heterogeneity.

\newcommand{\bestscore}[1]{\textbf{#1}}
\subsection{Ablation Study}

\begin{table}[t]
\centering
\footnotesize
\setlength{\tabcolsep}{2.2pt}
\renewcommand{\arraystretch}{1.0}

\begin{tabular}{@{}clcccc@{}}
\toprule
\multirow{2}{*}{\textbf{Dataset}}
& \multirow{2}{*}{\textbf{Variant}}
& \multicolumn{2}{c}{\textbf{Top 100 DEGs}}
& \multicolumn{2}{c}{\textbf{All Genes}} \\
\cmidrule(lr){3-4}
\cmidrule(lr){5-6}
&
& \shortstack{\textbf{C-DEGs}\\$\uparrow$}
& \shortstack{\textbf{W-Dist}\\$\downarrow$}
& \shortstack{\textbf{Centroid}\\\textbf{Acc} $\uparrow$}
& \shortstack{\textbf{PDS}\\$\uparrow$} \\
\midrule

\multirow{5}{*}{\textbf{Adamson}}
& w/o Decomp & 10.520 & 5.927 & 0.240 & 0.733 \\
& w/o Gene   & 8.773  & 5.772 & 0.267 & 0.787 \\
& w/o Sys    & 10.400 & 5.746 & 0.360 & 0.846 \\
& w/o GM     & \bestscore{11.107} & 5.781 & 0.280 & 0.797 \\
& \textbf{DeMixPert}
& 10.733
& \bestscore{5.481}
& \bestscore{0.560}
& \bestscore{0.947} \\
\midrule

\multirow{5}{*}{\textbf{Replogle}}
& w/o Decomp & 12.022 & 5.905 & 0.244 & 0.736 \\
& w/o Gene   & 9.733  & 5.852 & 0.333 & 0.760 \\
& w/o Sys    & 11.556 & 5.832 & 0.333 & 0.800 \\
& w/o GM     & \bestscore{14.356} & 5.833 & 0.289 & 0.773 \\
& \textbf{DeMixPert}
& 12.378
& \bestscore{5.614}
& \bestscore{0.778}
& \bestscore{0.938} \\

\bottomrule
\end{tabular}

\caption{The results of the ablation study.}
\label{tab2}
\end{table}

We conduct ablation experiments on Adamson and Replogle by comparing the complete model with four ablated variants. The w/o Decomp variant replaces decomposed response modeling with unified modeling. The w/o Gene and w/o Sys variants remove the embedding-derived perturbation-specific effect and the basal-state-dependent systematic response, respectively. For brevity, w/o GM denotes removal of the entire Gaussian prototype Invertible Network, including both the GM and the invertible network. The complete results are reported in Table~\ref{tab2}.

Table~\ref{tab2} shows that DeMixPert leads overall with the lowest W-Dist and highest Centroid Acc/PDS. Removing decomposition lowers Centroid Acc/PDS to 0.240/0.244 and 0.733/0.736 on Adamson/Replogle; removing the systematic component also degrades both, confirming complementary roles in perturbation specificity and transferable basal-state patterns. Removing the embedding-derived perturbation-specific effect reduces C-DEGs from 10.733/12.378 to 8.773/9.733, confirming its importance for unseen perturbations. Although w/o GM raises C-DEGs to 11.107/14.356, it worsens all other metrics, showing that the deterministic response center alone cannot recover population distributions.

\subsection{Sensitivity Analysis}
\begin{figure}[htb]
	\centering
	\includegraphics[width = 0.94\linewidth,height=0.15\textheight]{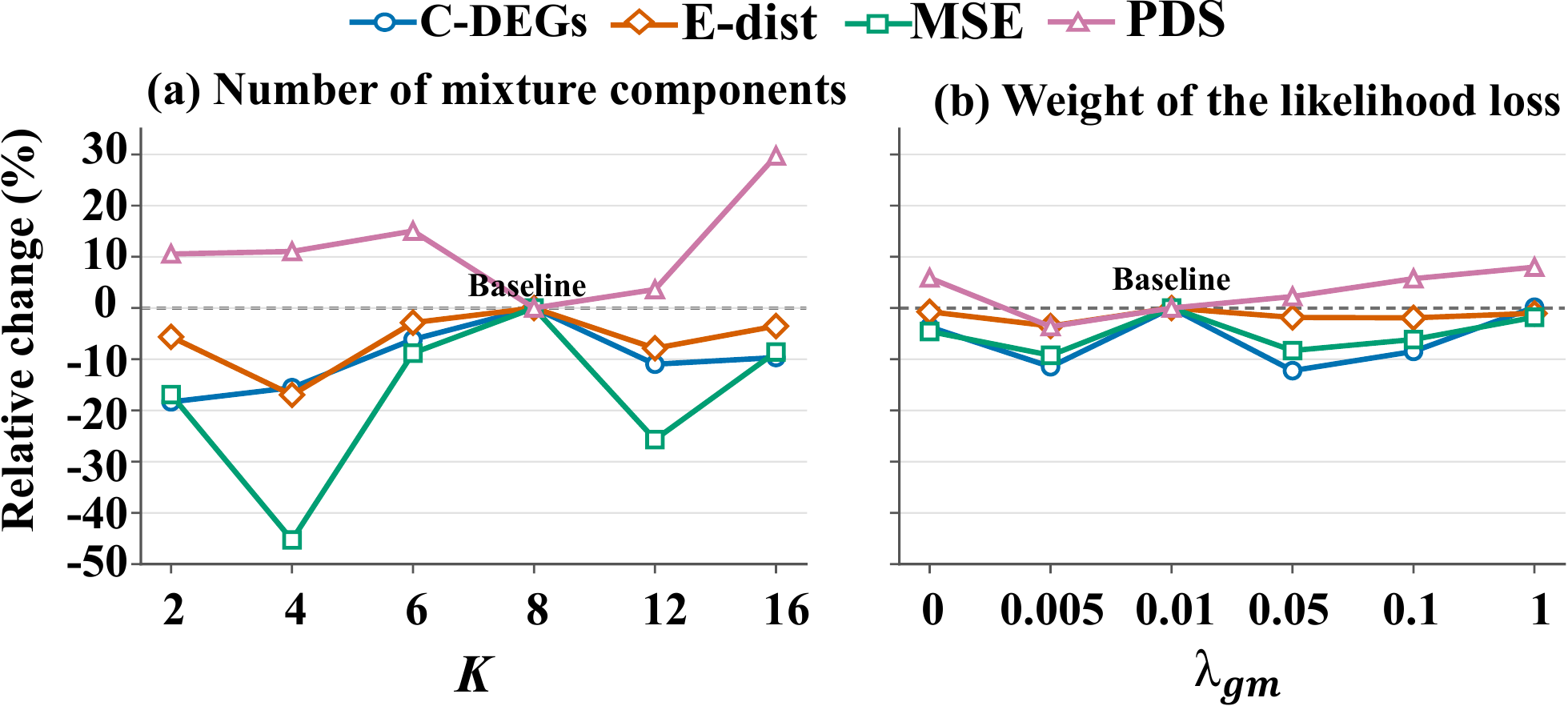}
\caption{The result of the sensitivity analysis. Positive values indicate improvement and negative values degradation.
}
\label{fig_sens}
\end{figure}

Fig.~\ref{fig_sens} shows performance changes under different Gaussian prototype numbers $K$ and GM-loss weights $\lambda_{\mathrm{gm}}$ relative to the reference setting ($K=8$, $\lambda_{\mathrm{gm}}=0.01$). Deviating from $K=8$ generally degrades DEG recovery, distribution modeling, and expression reconstruction, whereas larger $K$ improves perturbation identification, indicating a fidelity–separability trade-off. Similarly, increasing $\lambda_{\mathrm{gm}}$ improves identification but slightly affects other objectives. Overall, the reference setting achieves the best balance.

\begin{figure}[htb]
	\centering
	\includegraphics[width = 0.94\linewidth,height=0.30\textheight]{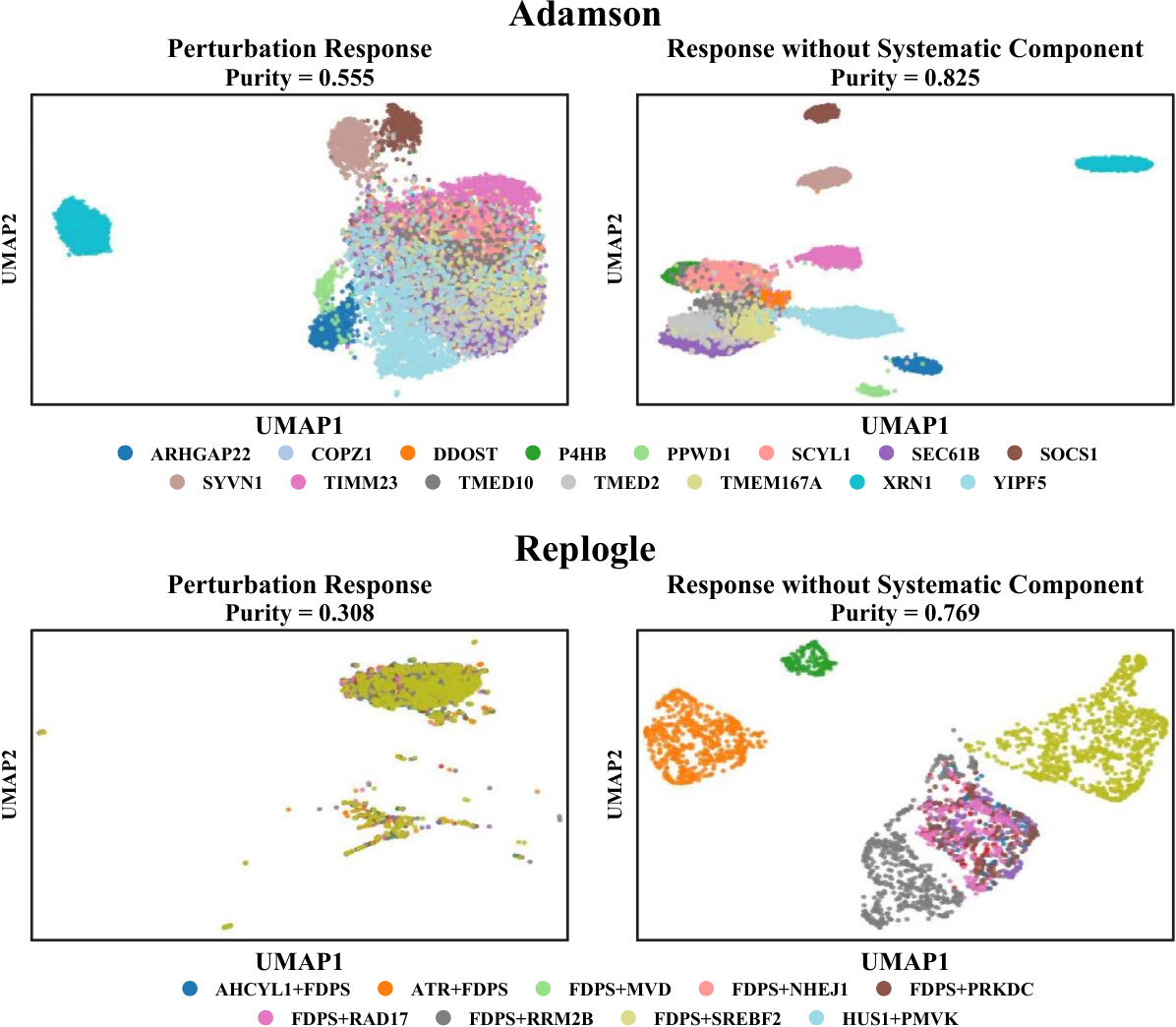}
\caption{UMAP visualizations of the perturbation response $r_i^*$ (left) and the response after removing the basal-state-dependent systematic component, $r_i^* - g_i$ (right), on the Adamson and Replogle datasets. Points are colored by perturbation condition, and cluster purity scores quantify the separation among perturbations.
}
	\label{fig5}
\end{figure}


\begin{figure}[htb]
	\centering
	\includegraphics[width = 0.97\linewidth,height=0.30\textheight]{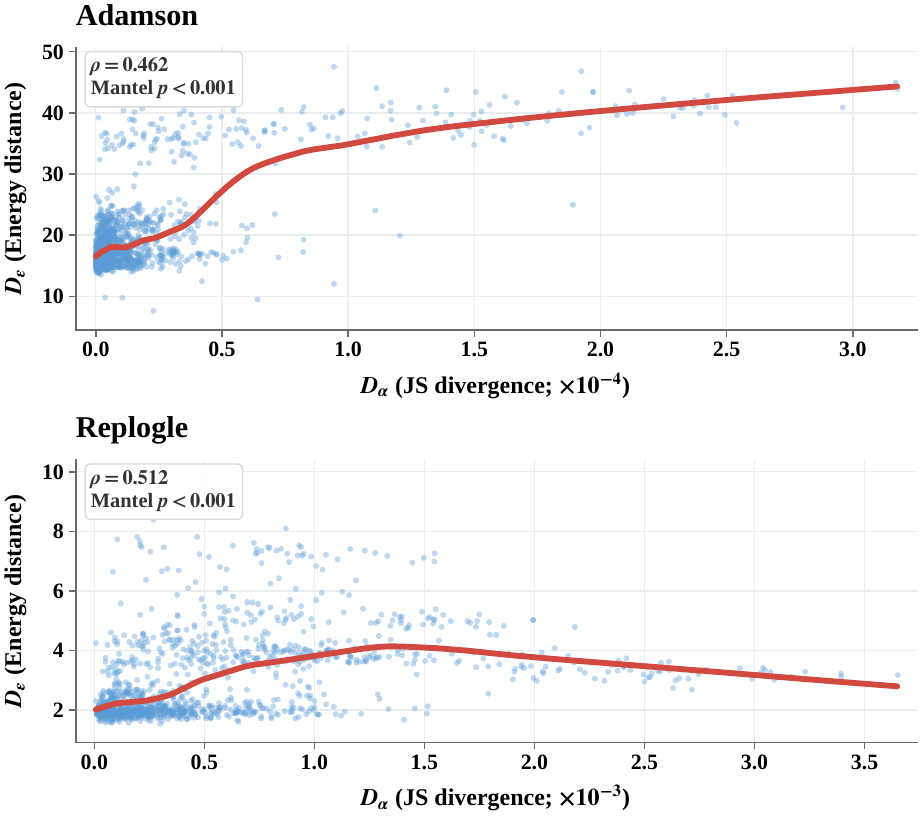}
	\caption{The correlation between Prototype-weight divergence and perturbation-specific population differences.
    Pairwise prototype-weight divergence
    $D_{\alpha}$ is compared with population-level distance
    $D_{\epsilon}$ in the Adamson  and Replogle
datasets. The red curves show the LOWESS-smoothed relationships
between $D_{\alpha}$ and $D_{\epsilon}$.}
	\label{fig6}
\end{figure}

\subsection{Interpretability Analysis}
Two interpretability analyses are conducted to validate two central design principles of DeMixPert: explicit response decomposition and condition-dependent composition of shared Gaussian prototypes.

Fig.~\ref{fig5} compares representations of predicted response  before and after removing the systematic component $g_i$. Cluster purity is applied to assess the separation of basal-state variation from perturbation-associated structure. 
It investigates whether the basal-state-dependent systematic response interferes with perturbation-associated structures in the predicted response space. After removing \(g_i\), the responses on both datasets form more compact and clearly separated clusters. Accordingly, cluster purity increases from \(0.308\) to \(0.769\) on Replogle and from \(0.555\) to \(0.825\) on Adamson. Since \(g_i\) is computed exclusively from the basal-state representation and is the only component removed in this comparison, these improvements indicate that it introduces variation that is not aligned with perturbation identity. This validates the response decomposition in DeMixPert and improves the discrimination of unseen perturbations.

Fig.~\ref{fig6} illustrates the relationship between Gaussian prototype weights and cellular distributions across different perturbation conditions. For each pair of perturbation conditions, $D_{\alpha}$ denotes the Jensen--Shannon (JS) divergence~\cite{lin2002divergence} between their Gaussian prototype weight vectors $\alpha$, whereas $D_{\epsilon}$ denotes the energy distance between their population distributions $\epsilon$. Spearman correlation is then used to assess the overall rank association between the two distance matrices~\cite{spearman1961proof}, and its statistical significance is evaluated using a Mantel-style permutation test~\cite{dietz1983permutation}.
Across all pairs of perturbation conditions, significant overall positive rank associations are observed in both the Adamson ($\rho=0.462$) and Replogle ($\rho=0.512$) datasets, with Mantel $p<0.001$ in both cases. The consistency of this association across the single-gene and combinatorial perturbation datasets indicates that differences in the learned prototype weights are closely associated with variation in perturbation-specific population distributions. These results provide empirical support for the effectiveness and interpretability of the Gaussian prototype composition mechanism.

\section{Conclusion}
In this study, we present DeMixPert, a Gaussian-mixture-based response decomposition framework for OOD single-cell perturbation prediction. DeMixPert disentangles perturbation responses into systematic, perturbation-specific, and population-level components, each modeled through a dedicated mechanism. A Gaussian prototype Invertible Network further characterizes cellular heterogeneity through context-adaptive prototype composition. Extensive experiments demonstrate accurate recovery of perturbation-specific responses under OOD settings. The adaptive prototype composition also captures condition-specific population structures, offering a structured interpretation of heterogeneous cellular responses.

\section*{Acknowledgements}
This work was supported in part by the National Key Research and Development Program of China (2024YFF1206600); in part by Guangdong S\&T Programme (2025B0101130001); in part by the National Natural Science Foundation of China (62325204, 62502161, 62502163); in part by the China Postdoctoral Science Foundation (2025M782912).

\clearpage
\appendix
\section{Supplementary Appendix}
\section{Perturbation Embedding Construction}
\label{app:perturbation_embedding}

As stated in the main paper, the perturbation embedding is constructed
from pretrained scGPT gene representations~\cite{cui2024scgpt}. Let
$\mathcal{M}_p$ denote the set of target genes named by perturbation
condition $p$, and let $\mathbf{e}_m\in\mathbb{R}^{512}$ be the
pretrained scGPT embedding of target gene $m$. We define
\begin{equation}
e_p
=
\frac{1}{|\mathcal{M}_p|}
\sum_{m\in\mathcal{M}_p}\mathbf{e}_m
\in\mathbb{R}^{512}.
\end{equation}
For a single-gene perturbation, this reduces to the embedding of that
gene. For a combinatorial perturbation, the target-gene embeddings are
averaged with equal weight. Conditions whose named target is unavailable
in the frozen embedding table are removed before split generation.

\begin{table*}[!t]
\centering
\scriptsize
\renewcommand{\arraystretch}{1.02}
\setlength{\tabcolsep}{6pt}

\begin{tabular}{@{}p{0.18\textwidth}p{0.76\textwidth}@{}}
\toprule
\textbf{Symbol} & \textbf{Definition} \\
\midrule

\multicolumn{2}{@{}l}{\textit{\textbf{Problem formulation}}} \\

$\mathcal{D}$ &
A single-cell genetic perturbation dataset. \\

$X^0$ &
The population of unperturbed control cells. \\

$X^p$ &
The observed cell population under perturbation condition $p$. \\

$x_j^0$, $x_i^p$ &
Gene-expression profiles of a control cell and a cell under
perturbation $p$, respectively. \\

$\mathcal{P}$, $p$ &
The set of perturbation conditions and an individual perturbation
condition, respectively. \\

$N_0$, $N_p$ &
The numbers of control cells and cells under perturbation $p$. \\

$G$ &
The number of selected genes in the expression space. \\

$\pi_0$ &
The distribution of unperturbed control cells. \\

$\pi(\cdot\mid p)$ &
The condition-specific distribution of cells under perturbation $p$. \\

$f_\theta$ &
The stochastic conditional generator parameterized by $\theta$. \\

$\Delta x_i^p$ &
The observed expression change defined in the main text,
$\Delta x_i^p=x_i^p-x_j^0$, where $x_j^0$ is a randomly sampled
control cell. \\

\addlinespace
\multicolumn{2}{@{}l}{\textit{\textbf{Response decomposition}}} \\

$E_s$, $E_r$ &
The basal-state encoder and response encoder, respectively. \\

$s_i\in\mathbb{R}^{d_s}$ &
The basal cellular-state representation of cell $i$. \\

$r_i^*\in\mathbb{R}^{d_r}$ &
The target response latent encoded from the observed expression
change $\Delta x_i^p$. \\

$d_s$, $d_r$ &
The dimensions of the basal-state latent and response latent spaces. \\

$g_i$ &
The basal-state-dependent systematic response component. \\

$e_{p_i}\in\mathbb{R}^{d_e}$ &
The pretrained embedding of perturbation condition $p_i$. \\

$d_e$ &
The dimension of the pretrained perturbation embedding. \\

$a_{p_i}$ &
The perturbation-level average response effect derived from
$e_{p_i}$. \\

$\bar{r}_i$ &
The deterministic response center,
$\bar{r}_i=g_i+a_{p_i}$. \\

$\epsilon_i^*$ &
The target cell-specific deviation,
$\epsilon_i^*=r_i^*-\bar{r}_i$. \\

\addlinespace
\multicolumn{2}{@{}l}{\textit{\textbf{Gaussian-prototype residual modeling}}} \\

$K$ &
The number of diagonal Gaussian response prototypes. \\

$\pi_k$, $\mu_k$, $\sigma_k^2$ &
The global mixture weight, mean, and diagonal variance of the
$k$-th Gaussian prototype. \\

$c_i$ &
The joint condition representation,
$c_i=\operatorname{concat}(s_i,a_{p_i})$. \\

$\alpha_{ik}$ &
The condition-dependent mixture weight of prototype $k$ for cell $i$. \\

$z_i$ &
A source variable sampled from the condition-dependent Gaussian
mixture. \\

$T_\theta$ &
The conditional invertible transformation that maps $z_i$ to a
cell-specific deviation. \\

$L$ &
The number of additive coupling layers in $T_\theta$. \\

$\gamma_{ik}^{(\ell)}$ &
The prototype-matching responsibility of prototype $k$ at coupling
layer $\ell$. \\

$\epsilon_i$ &
The generated stochastic cell-specific deviation. \\

$\hat{r}_i$ &
The predicted response latent,
$\hat r_i=\bar r_i+\epsilon_i$. \\

$D_\psi$ &
The expression decoder parameterized by $\psi$. \\

$\hat{x}_i^{p_i}$ &
The predicted post-perturbation expression profile. \\

\addlinespace
\multicolumn{2}{@{}l}{\textit{\textbf{Training objectives}}} \\

$\mathcal{L}_{\mathrm{align}}$ &
The response-latent alignment loss. \\

$\mathcal{L}_{\mathrm{gm}}$ &
The Gaussian-mixture negative log-likelihood loss. \\

$\mathcal{L}_{\mathrm{rec}}$ &
The expression reconstruction loss. \\

$\mathcal{L}_{\mathrm{dist}}$ &
The perturbation-condition-specific distribution matching loss. \\

$\mathcal{L}$ &
The overall training objective. \\

\bottomrule
\end{tabular}

\caption{Summary of the notation used in DeMixPert.}
\label{tab:notations}
\end{table*}

\section{Default Model Configuration}
\label{app:default_configuration}

This section reports the default architecture and optimization
configuration shared across all datasets. A summary of the model architecture is provided in
Table~\ref{tab:complete_architecture}, and dataset-specific training settings are reported in Supplementary
Appendix~H.

\begin{center}
\centering
\footnotesize
\renewcommand{\arraystretch}{1.12}
\begin{tabular}{@{}p{0.49\columnwidth}p{0.45\columnwidth}@{}}
\toprule
\textbf{Hyperparameter} & \textbf{Value} \\
\midrule
State latent dimension & 256 \\
Response latent dimension & 128 \\
Perturbation embedding dimension & 512 \\
Gaussian components $K$ & 8 \\
Coupling layers $L$ & 4 \\
Condition-network hidden width & 256 \\
Prototype temperature & 1.0 \\
Optimizer & AdamW \\
Learning rate & $5\times10^{-5}$ \\
Weight decay & $10^{-6}$ \\
Gradient clipping & global norm 5.0 \\
Learning-rate scheduler & none (constant learning rate) \\
\bottomrule
\end{tabular}
\captionof*{table}{Default configuration shared across datasets.
Dataset-specific batch sizes, evaluation intervals, Gaussian-mixture refresh
settings, and training schedules are listed in Supplementary Appendix~H.}
\end{center}

\begin{table*}[t]
\centering
\footnotesize
\renewcommand{\arraystretch}{1.12}
\setlength{\tabcolsep}{4pt}
\begin{tabular}{@{}p{0.16\textwidth}p{0.21\textwidth}p{0.17\textwidth}p{0.10\textwidth}p{0.15\textwidth}@{}}
\toprule
\textbf{Module} & \textbf{Layer dimensions} & \textbf{Activation} &
\textbf{Dropout} & \textbf{Norm.} \\
\midrule
State encoder & $G$--1024--512--256 & SiLU after hidden layers & 0.05 & LayerNorm after hidden linear layers \\
Response encoder & $G$--1024--512--128 & SiLU after hidden layers & 0.05 & LayerNorm after hidden linear layers \\
Systematic predictor & 256--512; $3\times$(512--1024--512); 512--128 & SiLU in blocks & 0.05 & LayerNorm at each block and output head \\
Anchor predictor & 512--512; $3\times$(512--1024--512); 512--128 & SiLU in blocks & 0.05 & LayerNorm at each block and output head \\
Condition network & 384--256--256; heads 8 and 32 & SiLU & 0 & None \\
Coupling flow & $4\times$ additive half-coupling; $8\rightarrow64$ shift map & None & 0 & None \\
Expression decoder & $384\rightarrow G$ & MLP-based mapping & -- & -- \\
\bottomrule
\end{tabular}
\caption{Main-text-aligned DeMixPert component summary. Hyphens
denote successive linear-layer widths; residual-block repetition is
written explicitly.}
\label{tab:complete_architecture}
\end{table*}

\section{Dataset Descriptions}
\label{app:datasets}

We evaluate DeMixPert on four single-cell genetic perturbation
datasets, comprising two single-gene perturbation datasets, Adamson
and Papalexi, and two combinatorial perturbation datasets, Norman and
Replogle. These datasets cover different perturbation technologies,
cellular contexts, and prediction settings. Dataset-specific
preprocessing and feature construction are described in
Supplementary Appendix~D, while the perturbation-level splitting
protocol is described in Supplementary Appendix~E.

\paragraph{Adamson.}

The Adamson dataset was generated using Perturb-seq in K562 cells to
characterize transcriptional responses associated with the unfolded
protein response~\cite{adamson2016multiplexed}. Each non-control
condition corresponds to a single targeted genetic perturbation.

\paragraph{Papalexi.}

The Papalexi dataset was generated using ECCITE-seq in THP-1 cells to
measure the transcriptional consequences of CRISPR perturbations
targeting immune-regulatory genes~\cite{papalexi2021}. The dataset
contains single-gene perturbation conditions.

\paragraph{Norman.}

The Norman dataset contains CRISPR activation measurements in K562
cells and was designed to study transcriptional responses and genetic
interactions induced by single-gene and combinatorial
perturbations~\cite{norman2019exploring}. The formal input uses one
shared gene space for all cells and perturbation conditions.

\paragraph{Replogle.}

The Replogle dataset contains both single-gene and dual-gene
perturbations measured using direct guide-RNA capture
technology~\cite{replogle2020}.

\begin{table*}[t]
\centering
\scriptsize
\renewcommand{\arraystretch}{1.12}
\setlength{\tabcolsep}{3pt}
\resizebox{\textwidth}{!}{%
\begin{tabular}{@{}p{0.09\textwidth}p{0.10\textwidth}p{0.08\textwidth}
p{0.10\textwidth}p{0.13\textwidth}p{0.11\textwidth}rrrrr@{}}
\toprule
\textbf{Dataset} & \textbf{Source} & \textbf{Species} &
\textbf{Cell line} & \textbf{Sequencing} & \textbf{Perturbation} &
\textbf{Cells} & \textbf{Control cells} & \textbf{Genes} &
\textbf{Single} & \textbf{Dual} \\
\midrule
Adamson & \cite{adamson2016multiplexed} & Human & K562 &
Perturb-seq & CRISPRi & 56,998 & 7,629 & 2,061 & 76 & 0 \\
Papalexi & \cite{papalexi2021} & Human & THP-1 &
ECCITE-seq & CRISPR knockout & 19,340 & 2,000 & 2,050 & 24 & 0 \\
Norman & \cite{norman2019exploring} & Human & K562 &
Perturb-seq & CRISPRa & 96,994 & 2,000 & 2,096 & 101 & 125 \\
Replogle & \cite{replogle2020} & Human & K562 &
direct-capture Perturb-seq & CRISPRi & 26,777 & 2,000 & 2,050 & 33 & 35 \\
\bottomrule
\end{tabular}
}
\caption{Dataset sources and post-filter statistics used in the
experiments. ``Single'' and ``Dual'' denote non-control perturbation
conditions; every dataset additionally contains one shared control
condition.}
\label{tab:dataset_statistics}
\end{table*}

\section{Data Preprocessing}
\label{app:preprocessing}

This section describes data quality control, normalization, gene
selection, perturbation embedding construction, and the response
reference used in the main-text formulation.

All models and metrics operate on the frozen \texttt{logNor} expression
layer. The feature space is the union of the 2,048 genes selected by
expression variability and the perturbation target genes that are
available in the dataset. The resulting gene counts are reported in
Table~\ref{tab:dataset_statistics}. Gene selection is performed once
when constructing each frozen input, before the
condition-level split is generated; it is therefore shared by all five
seeds and all compared methods.

Perturbation embeddings are constructed as defined in Supplementary
Appendix~A.

Following the main text, a control cell $x_j^0$ is randomly sampled
from the control population for each perturbed cell $x_i^p$, and the
response is defined as
\begin{equation}
\Delta x_i^p=x_i^p-x_j^0.
\end{equation}

\section{Out-of-Distribution Data Splitting Protocol}
\label{app:data_splitting}

We perform perturbation-level splitting to evaluate generalization to
unseen perturbation conditions. The non-control perturbations assigned
to the training, validation, and test sets are mutually disjoint.
Control cells are not treated as prediction targets in the split;
instead, they are retained as a shared reference population for
constructing perturbation responses and evaluating differential
expression.

\subsection{Single-Gene Perturbation Datasets}

For Adamson and Papalexi, all unique non-control perturbation
conditions are first sorted and then randomly permuted using a
seed-specific NumPy random number generator. Given $N$ non-control
conditions, the numbers assigned to the training and validation sets
are computed as

\begin{equation}
N_{\mathrm{train}}
=
\operatorname{round}(0.7N),
\qquad
N_{\mathrm{val}}
=
\operatorname{round}(0.1N),
\end{equation}

and the remaining conditions are assigned to the test set. For
datasets with at least three non-control conditions, the implementation
ensures that the validation and test sets are non-empty. The resulting
conditions within each split are stored in lexicographic order.

\subsection{Combinatorial Perturbation Datasets}

For Norman and Replogle, we follow the combinatorial perturbation
protocol of \cite{wei2026benchmarking}. All single-gene perturbation
conditions are included in the training set. The dual-gene conditions
are randomly permuted, and

\begin{equation}
N_{\mathrm{dual,train}}
=
\operatorname{round}
\left(
0.5N_{\mathrm{dual}}
\right)
\end{equation}

dual-gene conditions are assigned to training. If
$N_{\mathrm{remain}}$ denotes the number of remaining dual-gene
conditions, the validation set receives

\begin{equation}
N_{\mathrm{dual,val}}
=
\left\lfloor
\frac{N_{\mathrm{remain}}}{2}
\right\rfloor,
\end{equation}

and all other remaining dual-gene conditions are assigned to the test
set. Therefore, when the number of remaining dual-gene conditions is
odd, the test set contains one more condition than the validation set.

\subsection{Random Seeds and Split Validation}

All methods are evaluated using the same five data-splitting seeds:

\begin{equation}
\mathcal{S}
=
\{17, 23, 29, 31, 37\}.
\end{equation}

For each dataset and random seed, the generated split is saved before
model training and reused by DeMixPert and all comparative methods.
The implementation explicitly verifies

\begin{equation}
\mathcal{P}_{\mathrm{train}}
\cap
\mathcal{P}_{\mathrm{val}}
=
\mathcal{P}_{\mathrm{train}}
\cap
\mathcal{P}_{\mathrm{test}}
=
\mathcal{P}_{\mathrm{val}}
\cap
\mathcal{P}_{\mathrm{test}}
=
\emptyset.
\end{equation}

This condition-level separation ensures that the reported performance
measures generalization to unseen perturbation conditions rather than
memorization of cells from perturbations observed during training.

\section{Gaussian-Prototype EM Updates}
\label{app:gmm_updates}

This section describes the initialization, expectation-maximization
procedure, and periodic refresh strategy of the Gaussian response
prototypes.

\paragraph{Residual population.}
The Gaussian mixture model is fitted to training residuals only. With model parameters fixed
in evaluation mode, the implementation computes
\begin{equation}
\epsilon_i^*=E_r(\Delta x_i^p)-\bar r_i
\end{equation}
for every non-control training cell. At most the dataset-specific number
of residuals in Table~\ref{tab:dataset_hyperparameters} is sampled
uniformly without replacement. Residual extraction and mixture fitting use
\texttt{no\_grad}; consequently, the EM updates are not differentiated
through.

\paragraph{Initialization.}
For $K=8$, a seeded random permutation of the $N$ residuals is generated
and the first $K$ residual vectors initialize the component means. The
mixture weights are initialized uniformly, $\pi_k=1/K$. Every component
receives the same feature-wise population variance, with
$\epsilon_{\mathrm{cov}}=10^{-5}$:
\begin{equation}
\sigma_{kj}^{2(0)}
=
\max\!\left\{
\frac{1}{N}\sum_{i=1}^{N}
\left(\epsilon_{ij}^*-\bar\epsilon_j^*\right)^2,
\epsilon_{\mathrm{cov}}
\right\}.
\end{equation}
The covariance is diagonal; no full-covariance or tied-covariance
parameter is estimated.

\paragraph{E step.}
At EM iteration $t$, let
$v_{kj}^{(t)}=\max(\sigma_{kj}^{2(t)},\epsilon_{\mathrm{cov}})$.
Responsibilities are computed in log space:
\begin{align}
\ell_{ik}^{(t)}
&=
\log \max(\pi_k^{(t)},10^{-12})
\nonumber\\
&\quad-\frac{1}{2}\sum_{j=1}^{d_r}
\frac{(\epsilon_{ij}^*-\mu_{kj}^{(t)})^2}
{v_{kj}^{(t)}}
\nonumber\\
&\quad-\frac{1}{2}\sum_{j=1}^{d_r}
\left[\log v_{kj}^{(t)}+\log(2\pi)\right],
\\
\gamma_{ik}^{(t)}
&=
\exp\!\left[
\ell_{ik}^{(t)}
-\operatorname{LSE}_{m=1}^{K}\ell_{im}^{(t)}
\right],
\end{align}
where $\operatorname{LSE}$ denotes log-sum-exp.

\paragraph{M step.}
Let
$N_k^{(t)}=\max(\sum_i\gamma_{ik}^{(t)},10^{-8})$. The update is
\begin{align}
\pi_k^{(t+1)}
&=\frac{N_k^{(t)}}{N},
\\
\mu_k^{(t+1)}
&=\frac{1}{N_k^{(t)}}\sum_i\gamma_{ik}^{(t)}\epsilon_i^*,
\\
\sigma_{kj}^{2(t+1)}
&=
\max\!\left\{
\frac{1}{N_k^{(t)}}\sum_i\gamma_{ik}^{(t)}
\left(\epsilon_{ij}^*-\mu_{kj}^{(t+1)}\right)^2,
\epsilon_{\mathrm{cov}}
\right\}.
\end{align}
The variance floor is the covariance regularizer; no additional
Wishart, entropy, or Dirichlet penalty is used in EM.

\paragraph{Iteration and convergence.}
The mean log-likelihood computed in the E step,
\begin{equation}
\mathcal{Q}^{(t)}
=
\frac{1}{N}\sum_i
\operatorname{LSE}_{k=1}^{K}\ell_{ik}^{(t)}
\end{equation}
is monitored once per EM iteration. EM stops when
$|\mathcal{Q}^{(t)}-\mathcal{Q}^{(t-1)}|<10^{-4}$ or after the
dataset-specific maximum number of iterations in
Table~\ref{tab:dataset_hyperparameters}, whichever occurs first.

\paragraph{Empty and near-empty components.}
Responsibilities are soft, and the effective count is lower-bounded by
$10^{-8}$. Therefore, a near-empty component remains finite and is not
deleted, merged, or randomly reinitialized. If fewer than $K$ residual
samples are available, the refresh routine leaves the previous mixture
parameters unchanged rather than silently reducing $K$.

\paragraph{Refresh schedule and numerical stability.}
The Gaussian mixture model is initialized once at epoch 0 and refreshed after every
dataset-specific number of training epochs using newly recomputed
residuals. The refresh seed is the run seed at epoch 0 and
\texttt{run\_seed+epoch} thereafter. EM is performed in float32 on CPU
and the resulting weights, means, and variances replace non-trainable
model buffers. Besides the variance and effective-count floors, all
mixture normalization uses log-sum-exp, Gaussian log-probabilities use a
minimum variance of $10^{-6}$ at training/inference, and sampling applies
the same $10^{-6}$ floor before the square root. Conditional mixture
weights are produced with softmax; fixed global weights are clamped to
$10^{-8}$ before taking logarithms.

\section{Baseline References and Code}
\label{app:baselines}

Table~\ref{tab:baseline_links} lists the paper and official code
repository for each comparison method used in the main paper. We omit
repository-specific implementation details here and refer readers to
the corresponding public releases.

\begin{table*}[t]
\centering
\scriptsize
\renewcommand{\arraystretch}{1.12}
\setlength{\tabcolsep}{5pt}
\begin{tabular}{@{}p{0.10\textwidth}p{0.41\textwidth}p{0.41\textwidth}@{}}
\toprule
\textbf{Method} & \textbf{Paper} & \textbf{Official code} \\
\midrule
GEARS &
\url{https://www.nature.com/articles/s41587-023-01905-6} &
\url{https://github.com/snap-stanford/GEARS} \\
scGPT &
\url{https://www.nature.com/articles/s41592-024-02201-0} &
\url{https://github.com/bowang-lab/scGPT} \\
GenePert &
\url{https://www.biorxiv.org/content/10.1101/2024.10.27.620513v1} &
\url{https://github.com/zou-group/GenePert} \\
scDFM &
\url{https://openreview.net/forum?id=QSGanMEcUV} &
\url{https://github.com/AI4Science-WestlakeU/scDFM} \\
STATE &
\url{https://www.biorxiv.org/content/10.1101/2025.06.26.661135v1} &
\url{https://github.com/ArcInstitute/state} \\
\bottomrule
\end{tabular}
\caption{Paper and official code links for the five comparison
methods used in the main experiments.}
\label{tab:baseline_links}
\end{table*}

\section{Dataset-Specific Hyperparameters}
\label{app:dataset_hyperparameters}

This section reports the dataset-specific training, validation, and
Gaussian-prototype refresh settings used in the experiments.

\begin{table*}[t]
\centering
\scriptsize
\renewcommand{\arraystretch}{1.12}
\setlength{\tabcolsep}{3.3pt}
\resizebox{\textwidth}{!}{%
\begin{tabular}{@{}lrrrrrrrr@{}}
\toprule
\textbf{Dataset} & \textbf{LR} & \textbf{Cell batch} & \textbf{Max epochs} &
\textbf{Eval int.} & \textbf{EM int.} & \textbf{EM samples} &
\textbf{EM iter.} & \textbf{$K$} \\
\midrule
Adamson & $5\!\times\!10^{-5}$ & 1024 & 5000 & 1000 & 1000 & 5000 & 15 & 8 \\
Papalexi & $5\!\times\!10^{-5}$ & 4096 & 5000 & 100 & 100 & 50,000 & 50 & 8 \\
Norman & $5\!\times\!10^{-5}$ & 16,384 & 5000 & 200 & 200 & 20,000 & 20 & 8 \\
Replogle & $5\!\times\!10^{-5}$ & 1024 & 5000 & 200 & 100 & 30,000 & 50 & 8 \\
\bottomrule
\end{tabular}
}
\caption{Dataset-specific optimization and Gaussian-mixture refresh configuration.
``Cell batch'' is the number of non-control cells in one optimizer
update; gradient accumulation is not used.
``Int.'' is measured in training epochs.}
\label{tab:dataset_hyperparameters}
\end{table*}

\FloatBarrier

For every dataset,
$\lambda_{\mathrm{align}}=\lambda_{\mathrm{rec}}
=\lambda_{\mathrm{dist}}=1$ and
$\lambda_{\mathrm{gm}}=0.01$. The Gaussian mixture model is also
refreshed at epoch 0, so an EM interval of 100 denotes updates at epochs
$0,100,200,\ldots$.

\section{Additional Experimental Results}
\label{app:additional_results}

This section reports the sensitivity analysis of DeMixPert. The complete
ablation results are already reported in the main paper; the
module-level definitions and forward-path changes are provided in
Supplementary Appendix~J and are not duplicated as a second figure here.

\section{Ablation Details and Evaluation Metrics}
\label{app:metrics}

\subsection{Ablation Implementation Details}
\label{app:ablation_implementation}

The main paper defines four ablated variants:

\begin{itemize}
    \item \textbf{w/o Decomp} replaces the explicit three-part
    decomposition
    $\hat r_i=g(s_i)+a(e_{p_i})+\epsilon_i$ with a unified response
    mapping. The response is no longer represented by separately
    parameterized systematic, target-specific, and population components.

    \item \textbf{w/o Gene} removes the embedding-derived
    target-specific response by setting $a(e_{p_i})=\mathbf{0}$.
    Hence $\bar r_i=g(s_i)$; the pretrained embedding-derived
    deterministic response branch is absent.

    \item \textbf{w/o Sys} removes the basal-state-dependent systematic
    response by setting $g(s_i)=\mathbf{0}$, yielding
    $\bar r_i=a(e_{p_i})$. The state encoder remains active because
    $s_i$ is still used by the condition network and expression decoder.

    \item \textbf{w/o GM} removes the entire Gaussian-Prototype
    Invertible Network, including both the Gaussian mixture and the
    invertible network. No GMM is fitted, no EM refresh is performed,
    no coupling transform is applied, and
    $\epsilon_i=\mathbf{0}$. The prediction therefore reduces to the
    deterministic response center, $\hat r_i=g(s_i)+a(e_{p_i})$.
    The inapplicable likelihood term
    $\lambda_{\mathrm{gm}}\mathcal{L}_{\mathrm{gm}}$ is disabled.
\end{itemize}

The ablation comparison is conducted on Adamson and Replogle. Each
variant is independently initialized and retrained rather than produced
by post-hoc masking of a trained complete model. Architecture-specific
operations and their associated losses are disabled only when
inapplicable to the corresponding variant.

\subsection{Evaluation Protocol and Metric Definitions}
\label{app:evaluation_protocol}

We evaluate each test perturbation condition independently and then
macro-average the resulting scores across test conditions. Let
$\mathbf{X}^{p}\in\mathbb{R}^{N_p\times G}$ and
$\widehat{\mathbf{X}}^{p}\in\mathbb{R}^{\widehat{N}_p\times G}$
denote the observed and predicted expression matrices under
perturbation condition $p$, respectively. Their population-level
mean expression vectors are
\begin{equation}
\boldsymbol{\mu}^{p}
=
\frac{1}{N_p}\sum_{i=1}^{N_p}\mathbf{x}^{p}_{i},
\qquad
\widehat{\boldsymbol{\mu}}^{p}
=
\frac{1}{\widehat{N}_p}
\sum_{i=1}^{\widehat{N}_p}
\widehat{\mathbf{x}}^{p}_{i}.
\end{equation}
We denote the mean expression vector of the observed control
population by $\boldsymbol{\mu}^{0}$.

\subsubsection{Top-100 DEG Metrics}

\paragraph{Top-100 DEG selection.}
For each perturbation condition $p$, genes are ranked by comparing
the observed perturbed cells with the observed control cells using
the Wilcoxon rank-sum test. Let
$\mathcal{S}^{\mathrm{true}}_p$ denote the 100 highest-ranked genes.
The same procedure is applied to the predicted perturbed cells and
the observed control cells to obtain
$\mathcal{S}^{\mathrm{pred}}_p$.
The observed set $\mathcal{S}^{\mathrm{true}}_p$ is used as the
evaluation gene space for MSE, E-Dist, and W-Dist at the top-100 DEG
level.

\paragraph{Common differentially expressed genes (C-DEGs).}
C-DEGs measures the number of overlapping genes between the
top-100 DEG sets obtained from the observed and predicted
populations:
\begin{equation}
\mathrm{C\mbox{-}DEGs}_p
=
\left|
\mathcal{S}^{\mathrm{true}}_p
\cap
\mathcal{S}^{\mathrm{pred}}_p
\right|.
\end{equation}
The score ranges from $0$ to $100$, with a higher value indicating
better recovery of the perturbation-responsive genes. The values
reported in the main table may be non-integers because the
condition-level counts are macro-averaged.

\paragraph{Mean squared error (MSE).}
For a gene set $\mathcal{S}$, the population-level MSE is defined as
\begin{equation}
\mathrm{MSE}_p(\mathcal{S})
=
\frac{1}{|\mathcal{S}|}
\sum_{g\in\mathcal{S}}
\left(
\widehat{\mu}^{p}_{g}
-
\mu^{p}_{g}
\right)^2.
\end{equation}
At the top-100 DEG level, we use
$\mathcal{S}=\mathcal{S}^{\mathrm{true}}_p$. A lower value indicates
more accurate recovery of the mean perturbation response.

\paragraph{Energy distance (E-Dist).}
Energy distance evaluates the discrepancy between the complete
predicted and observed cell populations. Let
$\widehat{\mathbf{x}},\widehat{\mathbf{x}}'$ be independent samples
from the predicted distribution and
$\mathbf{x},\mathbf{x}'$ be independent samples from the observed
distribution. Energy distance is defined as
\begin{equation}
\begin{aligned}
\mathrm{E\mbox{-}Dist}_p
={}&
2\mathbb{E}
\left[
\left\|
\widehat{\mathbf{x}}-\mathbf{x}
\right\|_2
\right]
-
\mathbb{E}
\left[
\left\|
\widehat{\mathbf{x}}-\widehat{\mathbf{x}}'
\right\|_2
\right]
\\
&-
\mathbb{E}
\left[
\left\|
\mathbf{x}-\mathbf{x}'
\right\|_2
\right].
\end{aligned}
\end{equation}
For the top-100 DEG result, all expression vectors are restricted to
$\mathcal{S}^{\mathrm{true}}_p$. Lower E-Dist indicates better
agreement between the predicted and observed distributions. The code
uses the empirical V-statistic
\begin{equation}
2\overline{d}(\widehat X^p,X^p)
-\overline{d}(\widehat X^p,\widehat X^p)
-\overline{d}(X^p,X^p),
\end{equation}
where each bar is the mean of the complete pairwise Euclidean-distance
matrix, including its zero diagonal for within-population terms. The
result is clamped below at zero. Each population is subsampled without
replacement to at most 2,000 cells with a fixed condition-index seed.

\paragraph{Wasserstein distance (W-Dist).}
W-Dist measures the optimal-transport cost between the empirical
predicted and observed cell distributions. Using the squared
Euclidean ground cost, it is calculated as
\begin{equation}
\mathrm{W\mbox{-}Dist}_p
=
\left[
\sum_{i,j}
\gamma^{\star}_{ij}
\left\|
\widehat{\mathbf{x}}^{p}_{i}
-
\mathbf{x}^{p}_{j}
\right\|_2^2
\right]^{1/2},
\end{equation}
where $\gamma^{\star}$ denotes the transport coupling returned by
the shared optimal-transport implementation. W-Dist is evaluated on
$\mathcal{S}^{\mathrm{true}}_p$, and a lower value indicates greater
distributional similarity. The primary backend is
\texttt{Pertpy Distance} with the Wasserstein metric; no additional
cell subsampling is applied. If Pertpy is unavailable, the released
fallback computes an entropically regularized Sinkhorn cost with
$\varepsilon=0.08$ and 35 iterations and reports the square root of the
non-negative cost.

\subsubsection{All-Gene Metrics}

\paragraph{Differential Expression Score (DES).}
For each perturbation $p$, significant DEGs are identified separately
from the observed and predicted populations relative to the same
observed control population. We use the Wilcoxon rank-sum test with
Benjamini--Hochberg correction and an adjusted $p$-value threshold
of $0.05$. Let $\mathcal{D}^{\mathrm{true}}_p$ and
$\mathcal{D}^{\mathrm{pred}}_p$ denote the resulting significant gene
sets.

When the predicted set contains more genes than the observed set, it
is truncated to the
$|\mathcal{D}^{\mathrm{true}}_p|$ genes with the largest absolute
log-fold changes. Denoting the resulting predicted set by
$\widetilde{\mathcal{D}}^{\mathrm{pred}}_p$, DES is
\begin{equation}
\mathrm{DES}_p
=
\frac{
\left|
\mathcal{D}^{\mathrm{true}}_p
\cap
\widetilde{\mathcal{D}}^{\mathrm{pred}}_p
\right|
}{
\left|
\mathcal{D}^{\mathrm{true}}_p
\right|
}.
\end{equation}
Conditions with no significant observed DEGs are excluded from the
DES average. A higher DES indicates more accurate recovery of
statistically significant perturbation-responsive genes.

\paragraph{All-gene MSE.}
All-gene MSE uses the same definition as above, but is calculated over
the complete frozen gene space rather than
$\mathcal{S}^{\mathrm{true}}_p$.

\paragraph{Centroid accuracy.}
For every test perturbation, we compare its predicted population
centroid with the observed centroids of all test perturbations:
\begin{equation}
\mathrm{CA}_p
=
\mathbf{1}
\left[
\underset{q\in\mathcal{T}}{\arg\min}
\left\|
\widehat{\boldsymbol{\mu}}^{p}
-
\boldsymbol{\mu}^{q}
\right\|_2
=
p
\right],
\end{equation}
where $\mathcal{T}$ is the set of test perturbation conditions.
The final centroid accuracy is the mean of $\mathrm{CA}_p$ across
conditions. A higher value means that predicted populations preserve
condition-specific identities more accurately.

\paragraph{Perturbation Discrimination Score (PDS).}
The main comparison table denotes this metric as PDS. The L1-based
implementation below evaluates whether the predicted response of
condition $p$ is
most similar to the corresponding observed response rather than to
another test perturbation. We first define the predicted and observed
perturbation-effect vectors as
\begin{equation}
\widehat{\boldsymbol{\delta}}^{p}
=
\widehat{\boldsymbol{\mu}}^{p}
-
\boldsymbol{\mu}^{0},
\qquad
\boldsymbol{\delta}^{q}
=
\boldsymbol{\mu}^{q}
-
\boldsymbol{\mu}^{0}.
\end{equation}
For every pair $(p,q)$, the directly perturbed target genes of both
conditions are excluded, producing the comparison gene set
$\mathcal{G}_{p,q}$. The L1 distance is
\begin{equation}
d_{p,q}
=
\sum_{g\in\mathcal{G}_{p,q}}
\left|
\widehat{\delta}^{p}_{g}
-
\delta^{q}_{g}
\right|.
\end{equation}
Let $r_p$ be the rank of $d_{p,p}$ among
$\{d_{p,q}:q\in\mathcal{T}\}$ in ascending order. The condition-level
score is
\begin{equation}
\mathrm{PDS}_p
=
1-
\frac{r_p-1}{|\mathcal{T}|}.
\end{equation}
The final score is averaged across all test perturbations. A higher
PDS indicates better discrimination between different
perturbation responses. Ties are resolved deterministically by the
lexicographic condition name. If target-gene removal would eliminate
the entire feature space, the implementation falls back to all genes.
This L1-based quantity may be named \texttt{PDS-L1} in evaluator
artifacts, but it is reported as PDS in the main paper.

\paragraph{Aggregation protocol.}
Across the five seeds defined in Supplementary Appendix~E, the main
table reports the arithmetic mean and sample standard deviation
($\operatorname{ddof}=1$) across these five trials. DES conditions with
no significant observed genes are represented by NaN and omitted from
the DES macro-average. The per-condition evaluation caps are 64, 512,
256, and 256 cells for Adamson, Papalexi, Norman, and Replogle,
respectively. All methods are evaluated using the same frozen
evaluation implementation.

\section{Reproducibility Details}
\label{app:reproducibility}

\paragraph{Software environment.}
DeMixPert is implemented in Python and PyTorch. Data processing and
evaluation use AnnData, Scanpy, NumPy, pandas, and SciPy.

\paragraph{Random seeds.}
The five split seeds defined in Supplementary Appendix~E are also used
for model initialization. For every run, the Python, NumPy, PyTorch,
and CUDA random-number generators are initialized using the
corresponding seed.

\paragraph{Model selection.}
Model selection is performed exclusively on the validation set.
Specifically, the checkpoint with the lowest macro-averaged top-100
DEG MSE on the validation perturbation conditions is selected. The
test set is evaluated only after checkpoint selection and is never
used for model selection or hyperparameter tuning.
The same checkpoint-selection rule is applied to every random seed.

\section{Details of the Interpretability Experiments}
\label{app:interpretability}

\subsection{Separability of the Decomposed Response Representation}
\label{app:response-separability}

We first examined whether the response decomposition isolates
perturbation-specific information from variation driven by the basal cell
state. For a perturbed cell $i$, the response representation was obtained
from the expression difference between the perturbed cell $x_i$ and its
paired control cell $c_i$:

\begin{equation}
r_i^{*}
=
\operatorname{Enc}_{\mathrm{response}}(x_i-c_i).
\label{eq:raw-response}
\end{equation}

The model decomposes this representation as

\begin{equation}
r_i^{*}
\approx
g_i+a_p+\epsilon_i,
\label{eq:response-decomposition}
\end{equation}

where $g_i$ denotes the systematic response associated with the basal cell
state, $a_p$ is the perturbation-specific anchor for perturbation $p$, and
$\epsilon_i$ captures cell-level residual variation. The systematic component
is predicted from the latent state of the paired control cell:

\begin{equation}
s_i
=
\operatorname{Enc}_{\mathrm{state}}(c_i),
\qquad
g_i
=
f_{\mathrm{sys}}(s_i).
\label{eq:systematic-response}
\end{equation}

To assess the effect of removing state-driven variation, we compared the
following two response representations:

\begin{align}
r_i^{\mathrm{raw}}
&=
r_i^{*},
\label{eq:raw-representation}
\\
r_i^{\mathrm{specific}}
&=
r_i^{*}-g_i.
\label{eq:specific-representation}
\end{align}

The second representation removes the systematic component predicted from
the basal cell state while retaining the perturbation-specific anchor and
cell-level residual variation.

\paragraph{Experimental protocol.}

All analyses used the most recent baseline checkpoint selected according to
validation performance (\texttt{best.pt}). Checkpoints selected using test
performance (\texttt{best\_test.pt}), as well as checkpoints from sensitivity
or ablation experiments, were not used.

We evaluated the Adamson and Replogle datasets using five random seeds:

\begin{equation}
\mathcal{S}
=
\left\{17,23,29,31,37\right\}.
\end{equation}

For each dataset--seed combination, the analysis was restricted to
perturbations in the test split. To prevent perturbations with larger cell
populations from disproportionately influencing the clustering results, we
randomly sampled an equal number of cells from every test perturbation.

For every sampled cell, we computed the raw response latent $r_i^{*}$, the
control-state latent $s_i$, and the corresponding systematic response $g_i$.
The specific response was subsequently obtained as $r_i^{*}-g_i$.

Within each dataset--seed combination, the raw and specific response
representations were concatenated when fitting a \texttt{StandardScaler}.
The resulting common scaling transformation was then applied to both
representations. This ensured that the raw and specific responses were
evaluated using the same feature scaling.

K-means clustering was performed separately on the standardized raw and
specific representations. The number of clusters was set equal to the number
of held-out perturbations:

\begin{equation}
K
=
\#\{\text{test perturbations}\}.
\label{eq:number-clusters}
\end{equation}

The resulting cluster assignments were evaluated against the ground-truth
perturbation labels using the following metrics:

\begin{itemize}
    \item \textbf{Adjusted Rand Index (ARI):}
    agreement between the inferred clustering and the true perturbation
    partition, adjusted for chance.

    \item \textbf{Normalized Mutual Information (NMI):}
    normalized information shared by the inferred cluster assignments and
    the ground-truth perturbation labels.

    \item \textbf{Purity:}
    the proportion of correctly assigned cells when each cluster is labeled
    according to its most frequent ground-truth perturbation.

    \item \textbf{Silhouette score:}
    the compactness and separation of the inferred clusters in the response
    latent space.
\end{itemize}

All quantitative metrics were computed directly in the standardized,
high-dimensional response latent space. UMAP projections were used only for
visualization and were not used for clustering or metric calculation.

Higher ARI, NMI, purity, and silhouette scores for
$r_i^{\mathrm{specific}}$ than for $r_i^{\mathrm{raw}}$ indicate that
subtracting $g_i$ removes basal-state-driven variation and makes the
perturbation-specific response structure more separable.

\subsection{Association Between Prototype Weights and Residual Population Differences}
\label{app:prototype-residual-association}

We next investigated whether the Gaussian prototype weights learned by the
model reflect genuine differences between perturbation-specific residual
cell populations. Specifically, we tested whether perturbation pairs with
more dissimilar prototype-weight profiles also exhibit more dissimilar
residual-response distributions.

For perturbation $p$ under control-cell state $s_i$, the mixture weights over
the eight Gaussian prototypes are given by

\begin{equation}
\alpha_{i,p}
=
\operatorname{softmax}
\left(
h_{\mathrm{mix}}\left([s_i,a_p]\right)
\right),
\label{eq:prototype-weights}
\end{equation}

where $a_p$ is the perturbation embedding and
$\alpha_{i,p}\in\mathbb{R}^{8}$ contains the corresponding prototype
weights.

For each perturbation, we computed its average prototype-weight profile:

\begin{equation}
\bar{\alpha}_p
=
\frac{1}{M}
\sum_{i=1}^{M}\alpha_{i,p}
=
\frac{1}{M}
\sum_{i=1}^{M}
\operatorname{softmax}
\left(
h_{\mathrm{mix}}\left([s_i,a_p]\right)
\right),
\label{eq:mean-prototype-profile}
\end{equation}

where $M$ is the number of control cells used to calculate the profile.

Importantly, the same set of $M$ control cells was used for every
perturbation within a dataset--seed combination. This control ensures that
differences between the profiles $\bar{\alpha}_p$ are attributable to the
perturbations rather than to differences in basal cell-state composition.

\paragraph{Experimental protocol.}

We used the validation-selected baseline checkpoint (\texttt{best.pt}). The
main visualization was generated using the training split for Adamson seed
23 and Replogle seed 29.

For each pair of perturbations $(p,q)$, prototype-weight dissimilarity was
quantified using Jensen--Shannon divergence:

\begin{equation}
D_{\alpha}(p,q)
=
\operatorname{JS}
\left(
\bar{\alpha}_p,\bar{\alpha}_q
\right).
\label{eq:prototype-distance}
\end{equation}

We then derived the observed residual response for every perturbed cell. The
encoded response was

\begin{equation}
r_i^{*}
=
\operatorname{Enc}_{\mathrm{response}}(x_i-c_i),
\end{equation}

and the residual response was calculated by subtracting the systematic
response and perturbation anchor:

\begin{equation}
\epsilon_i^{*}
=
r_i^{*}
-
\left(g_i+a_p\right).
\label{eq:observed-residual}
\end{equation}

Let $E_p$ denote the collection of residual-response vectors associated with
perturbation $p$. To prevent differences in cell counts from biasing the
population-distance estimates, we sampled the same number of residual cells
for every perturbation. Specifically, we used 30 cells per perturbation for
Adamson seed 23 and 51 cells per perturbation for Replogle seed 29.

The balanced sampling procedure was independently repeated 20 times. For
each repetition $t$, the distance between the residual populations of
perturbations $p$ and $q$ was quantified using Energy Distance:

\begin{equation}
\operatorname{ED}
\left(
E_p^{(t)},E_q^{(t)}
\right).
\end{equation}

The final residual-population distance was calculated by averaging over the
20 sampling repetitions:

\begin{equation}
D_{\epsilon}(p,q)
=
\frac{1}{20}
\sum_{t=1}^{20}
\operatorname{ED}
\left(
E_p^{(t)},E_q^{(t)}
\right).
\label{eq:residual-distance}
\end{equation}

The Adamson training split contained 53 perturbations and therefore yielded

\begin{equation}
\binom{53}{2}
=
\frac{53\times52}{2}
=
1{,}378
\end{equation}

unique perturbation pairs. The Replogle training split contained 51
perturbations and yielded

\begin{equation}
\binom{51}{2}
=
\frac{51\times50}{2}
=
1{,}275
\end{equation}

unique pairs. Each perturbation pair contributed one observation of the form

\begin{equation}
\left(
D_{\alpha}(p,q),
D_{\epsilon}(p,q)
\right).
\end{equation}

We quantified the association between the two distance matrices by computing
Spearman's rank correlation over their upper-triangular elements:

\begin{equation}
\rho
=
\operatorname{Spearman}
\left(
\operatorname{upper}(D_{\alpha}),
\operatorname{upper}(D_{\epsilon})
\right).
\label{eq:distance-correlation}
\end{equation}

Because pairwise distances are not statistically independent, significance
was assessed using a Mantel-style permutation test rather than the
conventional $p$-value returned by a standard Spearman correlation test. In
each permutation, the perturbation labels of $D_{\alpha}$ were randomly
reordered by jointly permuting its rows and columns, while $D_{\epsilon}$ was
kept fixed. Spearman's $\rho$ was then recomputed. This procedure was
repeated 5,000 times, and the resulting permutation distribution was used to
calculate the empirical $p$-value.

The resulting associations were

\begin{align}
\text{Adamson, seed 23:}
\qquad
&\rho=0.462,
&
p_{\mathrm{Mantel}}<0.001,
\\
\text{Replogle, seed 29:}
\qquad
&\rho=0.512,
&
p_{\mathrm{Mantel}}<0.001.
\end{align}

\paragraph{Visualization.}

For each dataset, the prototype-weight heatmap displays perturbations as rows
and the eight Gaussian prototypes as columns. Each heatmap entry represents

\begin{equation}
\bar{\alpha}_{p,k},
\end{equation}

namely the average weight assigned by perturbation $p$ to Gaussian prototype
$k$. Perturbations were ordered by hierarchical clustering according to the
similarity of their prototype-weight profiles.

The distance-association panel displays $D_{\alpha}(p,q)$ against
$D_{\epsilon}(p,q)$ for all perturbation pairs. Because of the large number
of overlapping observations, the pairs were visualized using a
hexagonal-binning density plot. A LOWESS curve was added to show the overall
trend. This curve was used only as a visual guide and did not contribute to
the statistical significance test. The figure reports only Spearman's
$\rho$ and the Mantel permutation $p$-value.

The significant positive associations demonstrate that perturbations with
more dissimilar prototype-weight profiles also tend to have more dissimilar
residual cell populations. Thus, the learned Gaussian mixture weights are
not arbitrary internal coefficients but encode interpretable,
perturbation-specific population structure.

\FloatBarrier

\section{Construction of Sensitivity Scores}
\label{app:sensitivity_score_construction}

To summarize metrics with different scales and optimization
directions in the sensitivity analysis presented in the main paper, we convert each
metric into a direction-aligned percentage change relative to the
selected configuration, $K=8$ and $\lambda_{\mathrm{gm}}=0.01$.
Let $x_m(p)$ denote the five-seed mean of metric $m$ under parameter
setting $p$, and let $x_m^{\mathrm{ref}}$ denote its value under the
selected configuration. The transformed score is computed as
\begin{equation}
\Delta_m(p)=
\begin{cases}
100\left(\dfrac{x_m(p)}{x_m^{\mathrm{ref}}}-1\right),
& \text{if metric } m \text{ is higher-is-better}, \\[8pt]
100\left(1-\dfrac{x_m(p)}{x_m^{\mathrm{ref}}}\right),
& \text{if metric } m \text{ is lower-is-better}.
\end{cases}
\end{equation}
The higher-is-better metrics are C-DEGs, DES, Centroid Accuracy,
and PDS-$L_1$, whereas E-Dist, W-Dist, and MSE are
lower-is-better. We then obtain each curve by taking the unweighted
arithmetic mean of its constituent transformed metrics. Specifically,
DEG recovery averages C-DEGs and DES; Distribution averages top-100
E-Dist, top-100 W-Dist, and all-gene E-Dist; Expression error averages
top-100 MSE and all-gene MSE; and Perturbation ID averages Centroid
Accuracy and PDS-$L_1$. All-gene E-Dist is used only as an auxiliary
distributional measure in the sensitivity analysis and is not included
in the primary benchmark table. It is computed using the same
energy-distance definition over the complete input gene space.
Consequently, positive values indicate improvement over the selected
configuration, negative values indicate degradation, and the selected
configuration corresponds to $0\%$. During each sensitivity analysis,
one hyperparameter is varied while all remaining settings are held fixed.

\end{document}